%% file: main.tex
\documentclass[preprint, sort&compress]{elsarticle/elsarticle}

\usepackage{graphicx}
\usepackage{times}
\usepackage[utf8]{inputenc}
\usepackage{url}
\usepackage{amssymb}
\usepackage{amstext}
\usepackage{amsmath}
\usepackage{booktabs}
\usepackage{tabularx}
\usepackage{multirow, makecell} 
\usepackage{rotating}
\usepackage{nicefrac} 
\usepackage{setspace}
\usepackage{hyperref}
\usepackage{colortbl}
\usepackage{pgfplots}
\usepackage{comment}
\usepackage{threeparttable}
\usepackage{bm}
\usepackage{enumitem}
\usepackage{setspace}

\usepackage{chngcntr}
\counterwithin{table}{section} 

\usepackage{times}
\usepackage{balance}  
\usepackage{soul}               
\usepackage{url}
\usepackage{textcomp}
\usepackage{mathrsfs}

\usepackage[font=scriptsize]{subfig}

\usepackage{dcolumn}
\usepackage{color}
\usepackage{array}
 
\DeclareMathOperator*{\argmax}{arg\,max}

\newcommand{\bb}{\color{black}}

\usepackage{textcomp}
\usepackage{siunitx}

\usepackage{longtable}
\usepackage{pdflscape}

\usepackage{lineno}

\DeclareSIUnit{\day}{\text{\ensuremath{\mathrm{day}}}}
\begin{document}

\begin{frontmatter}

\title{
RadioPathomics: Multimodal Learning in Non-Small Cell Lung Cancer for Adaptive Radiotherapy}

\journal{arXiv}

\author[aff1]{Matteo Tortora\corref{cor1}}\ead{m.tortora@unicampus.it}

\author[aff1]{Ermanno Cordelli}\ead{e.cordelli@unicampus.it}

\author[aff1]{Rosa Sicilia}\ead{r.sicilia@unicampus.it}

\author[aff2]{Lorenzo Nibid}\ead{lorenzo.nibid@unicampus.it}

\author[aff3]{Edy Ippolito}\ead{e.ippolito@unicampus.it}

\author[aff2]{Giuseppe Perrone}\ead{g.perrone@unicampus.it}

\author[aff3]{Sara Ramella}\ead{s.ramella@unicampus.it}

\author[aff1]{Paolo Soda}\ead{p.soda@unicampus.it}

\address[aff1]{Unit of Computer Systems \& Bioinformatics \\ Department of Engineering, University Campus Bio-Medico of Rome\\  Via Alvaro del Portillo 21, 00128 Rome, Italy}

\address[aff2]{Anatomical Pathology \\ Department of Medicine, University Campus Bio-Medico of Rome\\  Via Alvaro del Portillo 21, 00128 Rome, Italy}

\address[aff3]{Radiation Oncology \\ Department of Medicine, University Campus Bio-Medico of Rome\\  Via Alvaro del Portillo 21, 00128 Rome, Italy}

\cortext[cor1]{Corresponding author: Matteo Tortora, E-mail: m.tortora@unicampus.it, tel: +39 06225419622, Address: Via Alvaro del Portillo 21, 00128, Rome, Italy}

\begin{abstract}
The current  cancer treatment practice collects  multimodal data, such as radiology images, histopathology slides, genomics and clinical data.
The importance of these data sources taken individually has fostered the recent raise of radiomics and pathomics, i.e.  the extraction of quantitative features from radiology and histopathology images routinely collected  to  predict clinical outcomes or to guide clinical decisions using artificial intelligence algorithms.
Nevertheless, how to combine them into a single multimodal framework is still an open issue.
In this work we therefore develop a multimodal late fusion  approach that combines hand-crafted features computed from radiomics, pathomics and clinical data to predict radiation therapy treatment outcomes for non-small-cell lung cancer patients.
Within this context, we investigate eight different late fusion rules (i.e. product, maximum, minimum, mean, decision template, Dempster-Shafer, majority voting, and confidence rule) and two patient-wise aggregation rules  leveraging the richness of information given by computer tomography images and  whole-slide scans.
The experiments in leave-one-patient-out cross-validation on an in-house cohort of 33 patients  show that the proposed multimodal paradigm  with an AUC equal to 90.9\% outperforms  each  unimodal approach,   suggesting that data integration can advance precision medicine.
As a further contribution,  we also compare the hand-crafted representations with features automatically computed by deep networks, and the late fusion paradigm with early fusion, another popular multimodal approach.
In both cases,  the experiments show that the proposed multimodal approach provides the best results.

\end{abstract}


\end{frontmatter}

\section{Introduction}
\label{sec:intro}
Nowadays, lung cancer is worldwide recognised as the most common type of cancer and one of the most frequent causes of tumour death despite the recent increase in the number of treatment options \cite{bray2018global}.
The current clinical decision-making process relies on multiple data sources to improve the detection and the classification as well as the prognosis of the tumours, such as radiology-based data (e.g. X-ray, CT scan, ultrasound, MRI and metabolic imaging), digital pathology slides, genome profiling and clinical data. 
Such a variety of modalities catch different clinical aspects of the cancer disease and can help clinicians to pursue the paradigm of precision medicine, i.e., tailoring the treatment to the specific patient.
Indeed, the wide variety of complementary quantitative bio-markers extracted from the various modalities can lead to more accurate diagnosis and more efficient treatment plans.

In the last decades, the artificial intelligence (AI) community has directed large efforts towards the detection and the classification of tumours using one or more modalities.
However, only in recent years we have assisted to a growing interest directed to the disease outcome prediction using different data modalities.
In this respect, the emerging areas of research are:
\begin{itemize}
\item \textbf{Genomics:} it is an interdisciplinary field of science that focuses on genomes, highlighting the role of human genetic variation in disease diagnosis, prognosis, and treatment response. 
However, genomics biomarkers still have limitations that hinder the possibility to collect such data in clinical routine due to their complexity and still high cost \cite{lee2017radiomics}.
\item \textbf{Radiomics:} it is based on the extraction of quantitative features from radiology images routinely collected in order to  predict clinical outcomes or guide clinical decisions using AI algorithms \cite{lambin2012radiomics, kumar2012radiomics}.
\item \textbf{Pathomics:} it refers to the combination of digital pathology, omic science and AI to extract embedded information in digitised high-resolution whole-slide images of tissue biopsy sections to obtain quantitative bio-markers~\cite{gupta2019emergence}.
\end{itemize}

Given the growing availability of public oncological datasets containing paired samples from different modalities, in the last few years researchers started to take into account the multimodal learning paradigm. 
Multimodal learning relies on the integration of heterogeneous data from multiple sources into a single machine learning framework.
Although several works use genomics, radiomics or pathomics data alone, there are still few works that aim to fuse these modalities together \cite{zhang2020novel,wu2021deepmmsa, chen2020pathomic, braman2021deep}.
In \cite{zhang2020novel} the authors proposed a multimodal radiomics model for preoperative prediction of lymphatic vascular infiltration in rectal cancer fusing features extracted from multiple radiology-based data sources (MR and CT).
Early results demonstrate performance improvement over the single modalities taken individually.
In \cite{wu2021deepmmsa} the authors proposed a multimodal deep learning method for non-small cell lung cancer (NSCLC) survival analysis fusing CT images in combination with clinical data. 
The preliminary results show that the proposed multimodal model improves the analysis of survival in NSCLC patients compared to the current state-of-the-art which only works with clinical data.
In \cite{chen2020pathomic} the authors proposed a deep multimodal fusion framework merging together histopathological images and genomics features to predict survival outcomes in patients with glioma or clear cell renal cell carcinoma.
Preliminary results show that the proposed multimodal fusion paradigm leads to an improvement over the current state-of-the-art in predicting survival outcomes when using each modality independently.
In~\cite{braman2021deep} the authors proposed a deep model merging together radiology scans, molecular profiling, histopathology slides and clinical factors to predict the overall survival of glioma patients.
Early results demonstrate that it significantly outperforms the best performing unimodal model.

The literature reported so far shows that few works have explored a multimodal approach, reporting performance improvement.
Furthermore, despite the importance of radiomics and pathomics taken individually, to the best of our knowledge  only one work to  date  has  combined  them  together into  a  single  machine learning framework~\cite{braman2021deep}.  
Hence, we present here another investigation that combines radiomics, pathomics and clinical data together into a single multimodal late fusion scheme to predict radiation therapy treatment outcomes for NSCLC patients.
In this work, we do not take into account genomics data, because in clinical practice, pending the results of ongoing studies, in patients with locally advanced NSCLC considered for chemoradiation treatment, knowledge of oncogene-dependent characteristics does not change the therapeutic strategy.
The late fusion scheme permits us to combine uncorrelated data flows that vary significantly in terms of dimensionality and sampling rates, as in our case.
To summarise, the AI-related novelties of this work are as follows:
\begin{itemize}
\item We proposed a multimodal late fusion scheme taking into account features extracted from radiomics, pathomics and clinical data.
\item The proposed approach shows as the simultaneous fusion of the three modalities leads to an improvement over the models fed with the stand-alone data flows.
\item We compare the proposed multimodal late fusion approach with the early fusion scheme and show how this latter has lower performance.
\item We also fed the multimodal learning scheme with deep features extracted from a pre-trained neural network fine-tuned with the modalities covered by this work.
Although, early results show as the deep features lead to performance deterioration, also in this case the combination of the three modalities leads to an improvement of the performances. 
\end{itemize}

The rest of this manuscript is organised as follows: section~\ref{sec:background} presents a short overview of the multimodal learning framework and its applications to oncology.
Section~\ref{sc:materials} introduces the materials, overviewing the multimodal data sources available.
Section~\ref{sc:method} presents the proposed multimodal learning framework, whilst section~\ref{sc:expresults} and section~\ref{sc:results} describe the experimental setup and the results respectively. 
Finally, section~\ref{sc:conclusion} provides concluding remarks.

\begin{figure}
    \centering
    \includegraphics[width=12 cm]{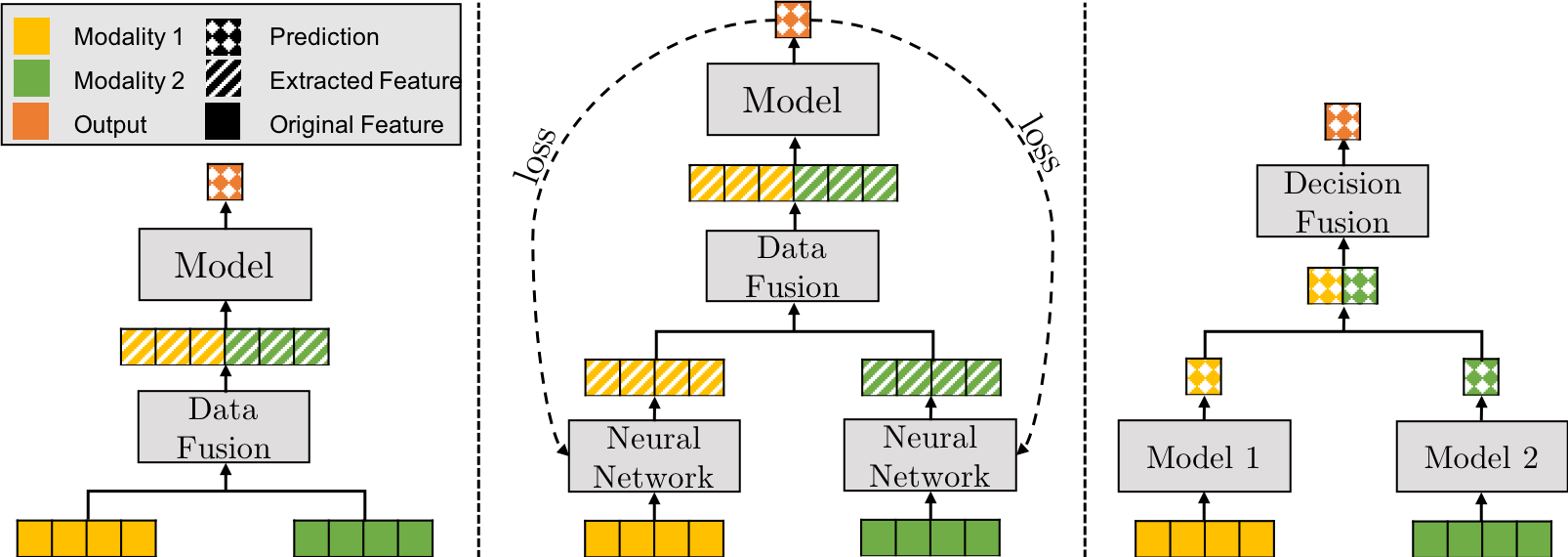}
    \caption{Model architectures for different multimodal learning frameworks. From left to right: early or data-level fusion, joint or intermediate fusion, late or decision-level fusion.}
    \label{fig:multimodal_framework}
\end{figure}

\section{Background}
\label{sec:background}
In this section, we first overview the various architectures in the multimodal learning framework, and then we summarise the current state-of-the-art on multimodal-based learning on oncology (section~\ref{subsec:multi-radiopath}).

\subsection{Multimodal learning}
\label{subsec:multi-learning}
Multimodal learning involves the integration of heterogeneous data from multiple sources extracted from the observation of the same phenomena or problem.
Hence, the use of multimodal data sources allows the extraction of a complementary, more robust and richer data representation, with the aim of improving performance compared to the use of a stand-alone modality.
Although there is not any formal proof, this intuition has brought interesting results in many applications, medical imaging included~\cite{huang2020fusion}.

Multimodal data integration can be performed at different levels using three types of fusion: early, joint, and late fusion~\cite{baltruvsaitis2018multimodal, ramachandram2017deep}, respectively (\autoref{fig:multimodal_framework}), as now described.

Early fusion, also known as data-level fusion or representation learning, concerns the integration of raw inputs from multiple data source modalities into a single feature vector before passing it into a single machine learning model (\autoref{fig:multimodal_framework}, left panel).
In the early fusion, raw input data can be merged into an  embedded space according to different policies, such as simple concatenation, addition, pooling, or applying a gated unit~\cite{kiela2018efficient}.
Although the promising results achieved in several applications, how to manage one or more missing modalities, how to handle different sampling rates and/or the time-synchronicity between multiple data sources, and the possible redundancies occurring while generating very large embedded spaces are the main issues of this multimodal approach.

With respect to joint and late fusion, these two combination schemes work by aggregating different classification models. Joint fusion, also known as intermediate-level fusion or hybrid fusion, concerns the combination of the extracted intermediate feature vectors from trained neural networks, one per modality, into an abstract fusion layer, also known as a shared representation layer (\autoref{fig:multimodal_framework}, central panel). 
Then, this combined feature vector feeds a final classification model whose loss is back-propagated to the feature extracting neural networks during training.
Since the loss is back-propagated during the training process, this fusion scheme improves the feature representation at each iteration leading to better multimodal embedded feature spaces.
Although joint learning is a very flexible framework, its main issue concerns the design of the architecture in terms of how, when, and which modalities can be fused~\cite{ramachandram2017deep}.

Let us now delve into late fusion approaches, as our proposed approach works at this level. 
In the early- to mid-2000s, late- or decision-level fusion has received considerable interest from the machine learning community due to its potential to improve the performance of stand-alone classifiers. 
Late fusion concerns the training of independent systems, one per modality, which are then combined by an aggregation function to reduce individual error rates (\autoref{fig:multimodal_framework}, right panel). 
This aggregation function takes as input the unimodal decision values provided by the different classifiers that are combined according to a fusion rule (e.g. minimum, maximum, mean, majority vote, etc.).
There is a consensus that the key to the success of late fusion is that it builds a mixture of diverse classifiers~\cite{cavalcanti2016combining}, providing different and complementary points of view to the ensemble.
Definitely,  the late fusion approach is a well-suited multimodal strategy when input modalities are significantly uncorrelated and they vary significantly in terms of data dimensionality and sampling rates~\cite{ramachandram2017deep}.
These are the major reasons that led us to explore this multimodal framework, which will be also experimentally compared against early fusion in section~\ref{sc:results}.

\subsection{Multimodal oncology}
\label{subsec:multi-radiopath}
Nowadays, the current clinical practice for cancer treatment requires collecting multimodal data for each patient, such as radiological images (e.g. X-ray, CT scan, ultrasound, MRI and metabolic imaging), histopathology slides, genomics and clinical data.
Such a variety of modalities describes different clinical aspects of cancer disease and can provide a wide range of complementary bio-markers leading to more accurate diagnosis and more efficient treatment plans.
Although there are several works in the current state-of-the-art dealing with the detection, classification and prognostic task taking the aforementioned single modalities individually \cite{demir2005automated, limkin2017promises, monkam2019detection, srinidhi2021deep}, there are still few works that aim to fuse these modalities together. 
Hence, in recent years, researchers focused their efforts on the fusion of these modalities into a single machine learning framework \cite{ zhang2020novel,wu2021deepmmsa, chen2020pathomic, braman2021deep}. 

In \cite{zhang2020novel} the authors proposed a novel multimodal radiomics model for preoperative prediction of lymphatic vascular infiltration (LVI) in rectal cancer based on hand-crafted features extracted from magnetic resonance (MR) and computed tomography (CT) modalities.
The authors validated their method on a retrospective cohort of 94 patients with histologically confirmed rectal cancer.
The early results show as the multimodal (MR/CT) radiomics models can serve as an effective visual prognostic tool for predicting LVI in rectal cancer. 
It demonstrated the great potential of preoperative prediction to improve treatment decisions over the stand-alone modalities.

In \cite{wu2021deepmmsa} the authors proposed a multimodal deep learning method for NSCLC survival analysis leveraging CT images in combination with clinical data. 
The authors validated their framework using data from The Cancer Imaging Archive (TCIA), which contains paired samples of CT scans and clinical data for 422 NSCLC patients~\cite{clark2013cancer}.
The preliminary results show that there is a relationship between prognostic information and radiomics images. 
In addition, the proposed multimodal model improves the analysis of survival in NSCLC patients compared to the current state-of-the-art which only works with clinical data.

In \cite{chen2020pathomic} the authors proposed a deep multimodal fusion framework for the end-to-end multimodal fusion of histopathological images and genomics features (mutations, CNV, mRNAseq) for survival outcome prediction. 
This work implements the Kronecker product to model pairwise feature interactions across modalities and controls the expressiveness of each modality through a gating-based attention mechanism. 
The authors validated their framework using glioma and clear cell renal cell carcinoma datasets from The Cancer Genome Atlas (TCGA), which contains paired samples of whole-slide images of hematoxylin-and-eosin-stained specimens, genotype, and transcriptome data for 769 patients~\cite{tomczak2015cancer}. 
Based on a 15-fold cross-validation, preliminary results show that the proposed multimodal fusion paradigm leads to an improvement over the current state-of-the-art in predicting survival outcomes when using  each modality independently.

In~\cite{braman2021deep} the authors proposed a deep model merging together radiology scans, molecular profiling, histopathology slides and clinical factors to predict the overall survival of glioma patients.
The authors validated their framework by collecting data from the TCIA repository, which contains paired samples of whole-slide images of hematoxylin-and-eosin-stained specimens, MRI scans, DNA sequencing data, and clinical variables for 176 glioma patients.
Early results show that their model, with a median C-index of \(0.788\pm 0.067\), significantly outperforms the best performing unimodal model, which has a median C-index equal to \(0.718\pm 0.064\).
Furthermore, the proposed model successfully stratifies patients into clinical subgroups based on overall survival, adding further granularity to clinical prognostic classification and molecular subtyping.

Despite the importance of radiomics and pathomics data taken individually, to the best of our knowledge, at the time this work was written, only one study has combined them into a single machine learning framework for the outcome prediction of radiation therapy treatment~\cite{braman2021deep}.
Hence, in this paper, we propose the second attempt to combine radiomics, pathomics and clinical data into a single model to predict outcomes for NSCLC patients.
As mentioned in \autoref{sec:intro}, since in the current clinical practice the knowledge of oncogene-dependent characteristics does not change the therapeutic strategy, in this work we do not consider genomics data

\begin{figure}
    \centering
    \includegraphics[width=12 cm]{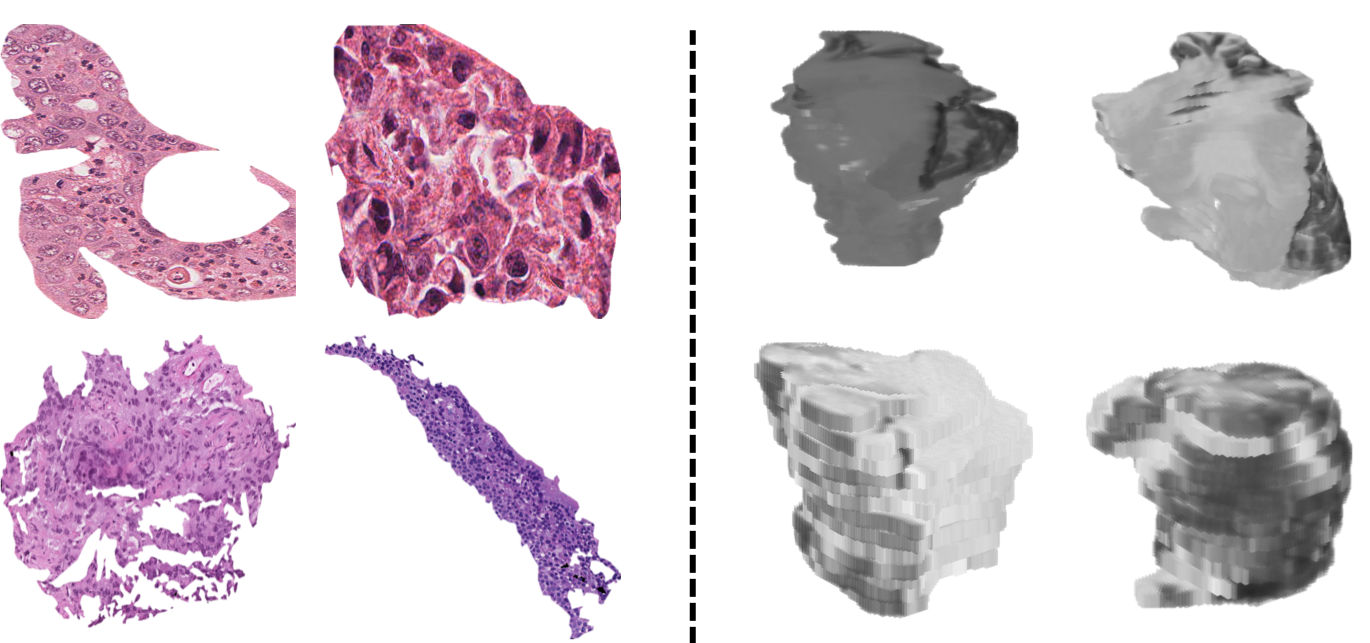}
    \caption{Example images for both pathomics and radiomics modality. To the left: examples of crops countered  by  pathologists. 
    To the right: examples of CTVs manually defined by expert lung pathology on CT scans.
    For the sake of presentation  we  show  both the  crops and the CTVs scaled  to  the  same size.}
    \label{fig:example-images}
\end{figure}
\section{Materials}
\label{sc:materials}
In this work we used an in-house cohort of 33 patients with Locally-Advanced stage III NSCLC, who were enrolled from November 2012 to July 2014 and treated with concurrent chemoradiation at a radical dose with an adaptive approach. 
The adaptive protocol was approved by Ethical Committee Campus Bio-Medico University on 30 October 2012 and registered at ClinicalTrials.gov on 12 July 2018 with Identifier NCT03583723 after an initial exploratory phase. 
Enrolled patients underwent a clinical evaluation after chemoradiation treatment and were classified into two groups according to target reduction: (i) adaptive, i.e. patients who achieved a reduction in tumour volume, assessed by two radiation oncologists on weekly chest CT simulations, leading to the implementation of a new treatment plan with which the patient would continue radiation therapy (adaptive approach); (ii) not-adaptive, i.e. patients who did not achieve target shrinkage and continued the chemoradiation with standard treatment.
The a priori probability of this patients' cohort consists of 11 and 22 adaptive and not-adaptive patients, respectively.

For this patient cohort we collected heterogeneous data including histological slides, CT scans, as well as clinical data, therefore forming the following unimodal data streams (i.e. pathomics, radiomics and semantics) in the multimodal learning framework investigated in this study:
\begin{itemize}
    \item \textbf{Pathomics modality}: This modality includes samples generated from biopsy slides of lung cancer tissue, stained with haematoxylin and eosin (HE). 
    HE (haematoxylin/eosin) tumour tissue slides were reviewed by a pathologist to confirm sample adequacy. Slides were digitised (APERIO CS2 Leica Biosystems or NanoZoomer 2.0 RT Hamamatsu) at 20x magnification.
    The digitised slides were loaded and segmented on QuPath. Regions of interest (ROIs), also called crops in the following, were manually defined by lung pathology experts to identify tumour areas avoiding histological artefacts, macrophage clusters and inflammations, fibrosis and necrosis.
    A total of 1113 tumour areas were manually segmented for the 33 tissue samples, one per patient.
    
    \item \textbf{Radiomics modality}: This modality includes initial CT scans collected prior to the start of concomitant chemoradiation therapy treatment.
    The CT scans consisted of single layer spiral computerised tomography - Siemens Somatom Emotion. 
    Acquisition parameters were 140 Kv, 80 mAs, and 3 mm for slice thick. 
    The scans were pre-processed applying a lung filter (kernel B70) and a mediastinum filter (kernel B31).
    The characteristics investigated in this work and presented in section~\ref{subsc:fea-extr} were extracted from 3D ROIs given by the Clinical Target Volume (CTV), manually countered by expert radiation oncologists.
    The CTV is the volume containing the Gross Tumour Volume (GTV), i.e. the macroscopically demonstrable disease, and therefore, with a probability considered relevant for therapy, the microscopic disease at the subclinical level.
    It is worth noting that in \cite{d2020radiomics} the authors showed that CTV should be preferred to GTV when computing radiomics features.
    In total, this modality contains 39 manually contoured CTVs for the 33 patients. It is worth noting that the number of CTVs exceeds the number of patients as multiple tumours can occur in a patient.
    
    \item \textbf{Semantic modality}: Two experienced radiation oncologists independently reviewed all CT scans and scored each tumour for four semantic imaging features, divided into tumour staging scores (T, N and tumour stage), and histological evaluation.
    They also added the age and sex of the patients.
    Each radiation oncologist blindly assigned staging scores, and, in case of disagreement, they reviewed the CT scans together and any discrepancies were resolved through discussion until consensus was reached.
\end{itemize}
As can be seen from these descriptions, the data sources used are highly heterogeneous and are uncorrelated unimodal flows. 
Thus, as we previously mentioned in \autoref{sec:intro}, this motivated the choice of using the late fusion approach as a multimodal approach. 

\autoref{fig:example-images} shows four examples of both crops extracted by the pathologists from the high-resolution whole-slide images which contain the selected tumour area of interest, and CTVs extracted by expert oncological radiotherapist by CT scans weekly collected during the radiation therapy treatment.
Moreover, the first three rows of \autoref{tb:cohort-distr} summarise the a priori sample distribution for each of the three different modalities.

\begin{table}[]
\centering
\begin{tabular}{lcccc}
\toprule
    &                \textbf{Modality}  & \textbf{Adaptive} & \textbf{Not-Adaptive } & \textbf{Total} \\
                   \midrule
\multirow{3}*{\textbf{Raw Data}} & Pathomics Crops & 303        & 810            & 1113    \\
& Radiomics CTVs & 13        & 26            & 39 \\ 
& Semantic & 11        & 22            & 33
\\ \midrule
\multirow{2}*{\textbf{\shortstack{Pre-Processed\\ Data}}} & Pathomics Patches & 10869        & 42681            & 53550    \\
& Radiomics Slices & 301        & 627            & 928 \\ 
\bottomrule
\end{tabular}
\caption{A priori distribution of samples for each unimodal flow.}
    \label{tb:cohort-distr}
\end{table}

\begin{figure}
    \centering
    \includegraphics[width=12.5 cm]{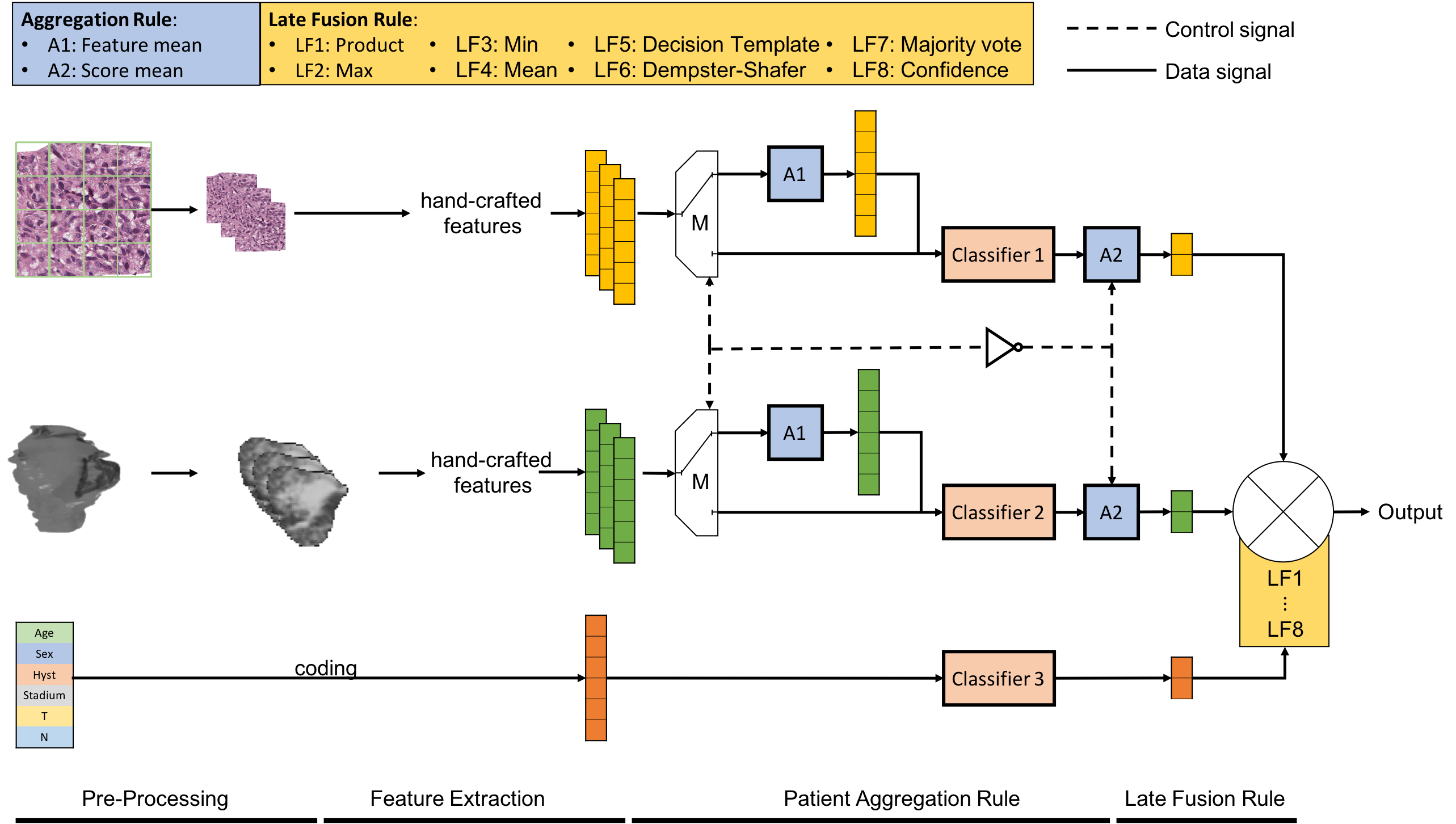}
    \caption{Proposed late fusion classification framework consisting of four blocks identified by the bars at the bottom of the figure. Let \emph{M} be a multiplexer that allows us switching between two patient aggregation modes, \(A_1\) and \(A_2\), respectively. If mode \(A_1\) is active, mode \(A_2\) is deactivated by a control signal passing through the logic NOT port, and vice versa.}
    \label{fig:framework}
\end{figure}

\section{Methods} 
\label{sc:method}
This section introduces the proposed fusion framework to handle the binary classification task introduced before. 
It is composed of four main blocks shown in~\autoref{fig:framework} identified by the bars at its bottom and presented in the following.
First, a pre-processing phase is applied to the different unimodal flows (section~\ref{subsc:pre-proc}). 
This stage uniforms the data, increases the dimensionality of unimodal flows, and encodes categorical features into numerical ones.
The second step, presented in section~\ref{subsc:fea-extr}, extracts the features from both images belonging to the pathomics and the radiomics flows.
The third step consists in patient aggregation, i.e. we merge each instance of the same patient of a single modality to get a single label for each patient (section~\ref{subsc:pat-agg}).
This is necessary so that the following data fusion step can work on consistent samples, i.e. one sample per patient, and not on single histologicals (patches or CTs' slices).
Section~\ref{subsc:fusion-rule} presents the eight fusion rules we investigated that belong to three different paradigms, this offering a view of how different fusion techniques can fuse three sources of information. 
On the one side, they are the \emph{product}, \emph{maximum}, \emph{minimum}, \emph{mean}, \emph{decision template} and \emph{Dempster-Shafer}, and all of them are based on the \emph{decision profile}, i.e., a matrix organising the output of \(l\) different soft classifiers in a multiple classier framework, with \(l=3\) in our case.
On the other side, the other two are the \emph{majority voting rule} that works with the crisp labels and the \emph{confidence rule}, which is a classifier selection technique.

\subsection{Pre-Processing}
\label{subsc:pre-proc}
This section presents the pre-processing applied to each unimodal flow which differs for each modality due to the heterogeneity of the data. 

In the case of pathomics images, we applied a patch extraction operation to the original crops manually contoured by expert pathologists on the high-resolution whole-slide images.
The patch extraction phase was performed by a sliding window with a size equal to \(100\times 100\) and a stride equal to 60, both chosen empirically. 
This step permits us to increase the cardinality of the available images by exploiting the variability typical of different regions of the same image. 
Furthermore, it also provides homogeneous images, as the original ones are characterised by a wide size variability. 
After this operation, we empirically removed the extracted patches with more than \(20\%\) of pixels belonging to the background to keep only the most informative images. 
In the end, the new repository of pathomics patches is composed of 53550 instances.

Let us now focus on the pre-processing for radiomics.
As already reported in section~\ref{sc:materials}, the radiomics modality consists of CTVs manually contoured by expert radiation oncologists on CT scans. 
To increase the dimensionality of this modality, we decomposed the CTVs,  i.e. the volumes containing the macroscopically demonstrable tumour mass and the microscopic disease at the sub-clinical level, into their component slices.
Thus we passed from having few 3D CTVs per patient to multiple 2D slices per patient, for a total of 928 slices.

Finally, in the case of semantic information, in order to have numerical features, we applied an ordinal encoding for T, N and stadium features, and one-hot encoding for sex and diagnosis features as no ordinal relation exists for these latter categorical variables.

The second part of \autoref{tb:cohort-distr} shows the a priori distribution of instances obtained after the pre-processing.

\subsection{Features Extraction}
\label{subsc:fea-extr}
This section describes the features extraction stage implemented for both the pathomics and the radiomics unimodal flows. 
It is straightforward that this step is not applied to the semantic unimodal flow, since it includes patients' medical data already processed.
Note also that the features computed for each modality were selected to optimise each unimodal flow, but this is out of the scope of this work and, for the sake of brevity, we do not present this phase here. 
Nevertheless, the starting feature set consists of 2D intensity and texture features well established in the medical image processing scenario \cite{chaddad2015radiomics} and, specifically, in radiomics \cite{aggarwal2012first} and digital pathology \cite{rashmi2020comparative}. 
They are statistical features extracted from the first-order image histogram, and several descriptors extracted from the results provided by both the Grey Level Co-Occurrence Matrix (GLCM) and the Local Binary Pattern (LBP) operators.
Please note that we also investigated the use of deep learning, as we will discuss in section~\ref{sc:expresults}.

\subsubsection{Pathomics}
Among the descriptors mentioned before, histopathological images are represented by measures derived from the GLCM, computed from each patch\footnote{Note that the GLCM operator is applied on the S-channel of the HSV colour model.}. 
The GLCM is a texture feature used to analyze the spatial distribution of grey levels within a 2D image at the microscale~\cite{haralick1973textural}.
Hence, given a \(n\times m\) image \(I\), and being \(l\) the number of bits used to represent it, a GLCM is a square matrix of size \(2^{l}\) where each entry \((g_i, g_j)\) represents the number of times a pixel with intensity \(g_i\) is separated from another pixel with intensity \(g_j\) by spatial offsets \(\Delta x\) and \(\Delta y\), respectively, in the vertical and horizontal directions: 
\begin{equation}
    { G }_{ \Delta x,\Delta y }\left( g_i,g_j \right) =\sum _{ x=1 }^{ n }{ \sum _{ y=1 }^{ m }{ \begin{cases} 1, & \text{if \(I\left( x,y \right) =g_i\) and \(I\left( x+\Delta x,y+\Delta y \right) =g_j\)}  \\ 
    0, & \text{Otherwise} \end{cases} }  },
\end{equation}
where \(g_i\) and \(g_j\) are the pixel values \(\in \left[ 0,{2}^{ l }-1 \right] \), \(x\) and \(y\) are the spatial positions in the image \(I\).

The GLCM can also be parameterised in terms of \(\delta\) and \(\theta\) instead of the spatial offsets (\(\Delta x,\Delta y\)).
The former represents the relative distance in pixels between two points in \(I\), while the latter is the relative orientation between two points in \(I\).
Here, taking into account preliminary experiments and findings in related fields~\cite{kral2016lbp,cordelli2018decision,sicilia2019early,joo2021stability}, for pathomics we used \(\delta=1\) and \(\theta\in[\ang{0}, \ang{45}, \ang{90}, \ang{135}] \), so that each patch has four GLCMs.

From each of these matrices we extracted six Haralick descriptors \cite{haralick1973textural}, i.e. contrast, dissimilarity, homogeneity, energy, correlation and angular second moment, listed in depth in Appendix B.
Their concatenation provides \(24\) textural descriptors per patch.

\subsubsection{Radiomics}
For the radiomics modality we used the same features as presented in our previous work \cite{ramella2018radiomic}. 
Hence, for each slice that makes up the 3D ROI extracted from the CT scans by the radiologists, we computed 12 statistics features and 104 textural features.

Statistical features consist of the moments up to the fourth-order of the first-order image histogram, i.e., the mean, the standard deviation, the skewness and the kurtosis. 
Furthermore, the picture of grey-level distribution is also grasped by the histogram width, the energy, the entropy, the value of the histogram absolute maximum and the corresponding grey-level value, the energy around such maximum, the number of relative maxima in the histogram and their energy.
These statistics are listed in depth in Appendix B.

Texture features are derived from the GLCM and from the LBP.
The former is parameterised by a unit distance \(\delta\) between pixels and an orientation \(\theta\in[\ang{0}, \ang{45}, \ang{90}, \ang{135}] \) and we extracted six second-order statistical features as the pathomics flow.

The latter is an operator that describes the local texture of an image by assigning each pixel an integer label according to its local circular neighbourhoods of \(P\) points located on the circumference  of radius \(R\) centred at  pixel \(\left( { x }_{ c },{ y }_{ c } \right) \)~\cite{ojala1996comparative}.
In this work, we used the  \({ LBP }_{ P,R }^{ riu2 }\)  operator, which is an extension of the original operator making it invariant to both local monotonic greyscale variations and rotation~\cite{ojala2002multiresolution}:
\begin{equation}
    { LBP }_{ P,R }^{ riu2 }\left( { x }_{ c },{ y }_{ c } \right)=\begin{cases} \sum _{ p=0 }^{ P-1 }{ sgn\left( { g }_{ p }-{ g }_{ c } \right)  },  & \text{if \(U\left( { LBP }_{ P,R } \right) \le 2\)}  \\ P+1, & \text{Otherwise} \end{cases}
\end{equation}
where \({ LBP }_{ P,R }\left( { x }_{ c },{ y }_{ c } \right) =\sum _{ p=0 }^{ P-1 }{ sgn\left( { g }_{ p }-{ g }_{ c } \right) { 2 }^{ p } } \) is the original LBP operator, \(sgn\left( \cdot  \right) \) is the sign function, \(g_c\) and \(g_p\) are the values of the grey intensity of the central pixel \(\left( { x }_{ c },{ y }_{ c } \right) \) and of the \(p\)-th pixel in the considered neighbourhood, respectively, and where \(U\left(\cdot \right)\) is the uniformity degree of a pattern:
\begin{equation}
    \begin{split}
        U\left( { LBP }_{ P,R } \right) &=\left| sgn\left( { g }_{ P-1 }-{ g }_{ c } \right) -sgn\left( { g }_{ 0 }-{ g }_{ c } \right)  \right| + \\
        &+\sum _{ p=1 }^{ P-1 }{ \left| sgn\left( { g }_{ p }-{ g }_{ c } \right) -sgn\left( { g }_{ p-1 }-{ g }_{ c } \right)  \right|  } 
    \end{split}    
\end{equation}
Then, this operator identifies \(P+1\) uniform patterns with decimal code in the range \(\left[ 0, P \right] \), whilst all non-uniform patterns are grouped into a single code with an integer value of \(P + 1\).
Hence, \({ LBP }_{ P,R }^{ riu2 }\) identifies a total of \(P + 2\) different decimal codes.
Finally, once we have applied this operator to each pixel in the image, we can compute a histogram of the LBP decimal codes' occurrences. 

In this work, we empirically parameterised \(R\) with a unit distance and we set \(P\) equal to \(8\). 
Finally, the same 12 statistical features reported above on the top of this section (i.e. mean, standard deviation, skewness, kurtosis, etc.) are then computed from the histogram of LBP distribution.
\bb

\subsection{Patient Aggregation Rule}
\label{subsc:pat-agg}
As mentioned above, each patient is composed of several samples both for the pathomics and the radiomics flows. 
For the former, the samples correspond to the patches extracted from the crops contoured by the pathologists, whilst for the latter, the samples correspond to the various slices included in the segmented CT VOIs.
 
For this reason, in order to have a single classification per patient and consistent sample fusion, a samples' patient-wise aggregation is necessary. 
In this work, we used two different patient-wise aggregation rules, denoted as \(A_1\) and \(A_2\) in~\autoref{fig:framework}, respectively. 

The former is applied before the classification step, and it averages out each component of the feature vector \(\bm{x} \in { \Re  }^{ n }\) belonging to the same patient $p$:
\begin{equation}
    { { \bm{x} }^{ p } }=\frac { 1 }{ { N }^{ p } } \sum _{ \bm{x}\in \mathcal{X}^{p}  }{ { \bm{x} } } 
\end{equation}
where \(\mathcal{X}^{p}\) is the set of feature vectors computed from all the samples of the same patient $p$ (i.e. histopathology patches or CT slices), and \(N^{p}\) is its cardinality.

The latter works after the classification process, and it averages the soft labels of all the instances of a patient.
Formally, given a classification problem with \(C\) class labels and \(L\) unimodal flows\footnote{Note that in our case \(C=2\) and \(L=3\).}, and assuming one classifier per modality, let \(\mathcal{D}=\left\{ { D }_{ i }\right\}_{i=1}^{L} \) denotes the set of classifiers. 
Hence, given \(\bm{x} \), a soft classifier outputs  a \(C\)-dimensional vector given by
\begin{equation}
    { D }_{ i }\left( \bm{x} \right) ={ \left[ { d }_{ i,1 }\left( \bm{x} \right) ,\dots ,{ d }_{ i,C }\left( \bm{x} \right)  \right]  }^{ T },
\end{equation}
where \({ d }_{ i,j }\left( \bm{x} \right) \in \left[ 0,1 \right] \) is the soft label and it  represents the degree of support provided by classifier \(D_{i}\) for the hypothesis that \(\bm{x}\) comes from the class \(\omega_{j}\).
On this premise, the \(A2\) patient-wise aggregation rule is defined by:
\begin{equation}
      { { d^p }_{ i,j } }=\frac { 1 }{ { N }_{ p } } \sum _{ \bm{x}\in { \mathcal{X}^{p} } }^{  }{ { d_{i,j} \left( \bm{x} \right)  } } 
\end{equation}
where, thus,  $d_{ i,j }^p$ represents the average soft label per class computed over all the instances of the $i$-th modality of the same patient (i.e. \(\bm{x} \in \mathcal{X}^p\)).

\subsection{Late Fusion Rules}
\label{subsc:fusion-rule}
This section introduces the late fusion rules that merge the multimodal information extracted from the different unimodal flows.

Using the notation already introduced, in a multimodality  framework we organise the   outputs returned by the $L$ unimodal classifiers   into a patient-wise \emph{decision profile} $DP^p$, defined by the following  matrix: 
\begin{equation}
     DP^p   =\begin{bmatrix} { \mu }_{ 1,1 }  & \cdots  & { \mu }_{ 1,j }  & \cdots  & { \mu }_{ 1,C }  \\ \vdots  & \ddots  & \vdots  &  & \vdots  \\ { \mu }_{ i,1 }  & \cdots  & { \mu }_{ i,j }  & \cdots  & { \mu }_{ i,C }  \\ \vdots  &  & \vdots  & \ddots  & \vdots  \\ { \mu }_{ L,1 }  & \cdots  & { \mu }_{ L,j }  & \cdots  & { \mu }_{ L,C }  \end{bmatrix}
\end{equation}
where $\mu_{i,j}$ is  computed according to $A1$ or $A2$ aggregation rule.
This implies that, using $A1$, $\mu_{i,j} = d_{i,j} (\bm{x}^p)$, whereas using $A2$ we have $\mu_{i,j} = d_{i,j}^p$.
Thus, the   patient-wise data is projected into a new feature space with dimension \(L\times C\) and this new representation combining the unimodal classification stages is depicted by the symbol \(\bigotimes \) in~\autoref{fig:framework}. 

The fusion methods calculate the support \(\chi_{j}\) for the class \(\omega_j\) by applying some mathematical procedure described below on the  \(DP^p\) representation and, using the maximum membership rule, we then assign the patient $p$ to the class \(\omega_s\) if:
\begin{equation}
    { \chi  }_{ s } \ge { \chi  }_{ z }  , \quad \forall \,\,\, z=1,\dots ,C
\end{equation}

In this work, to compute $\chi_j$ we apply eight  late fusion techniques, represented with the tag \(LF_t\), with $t = 1, \dots, 8$, in~\autoref{fig:framework}. 
They include four fusion rules computing the support for the \(j\)-th class independently of the support of the other classes:
\begin{enumerate}
    \item \textbf{Product rule (\(\bm{LF_1}\))}: it computes the support \(\chi_{j}\) for the class \(\omega_j\) as:
    \begin{equation}
        { \chi  }_{ j } =\prod _{ i=1 }^{ L }{ { \mu }_{ i,j } } 
    \end{equation}
    \item \textbf{Max rule (\(\bm{LF_2}\))}: it computes the support \(\chi_{j}\) for the class \(\omega_j\) as:
    \begin{equation}
        { \chi  }_{ j } =\max _{ i }{ { \mu }_{ i,j } } 
    \end{equation}
    \item \textbf{Min rule (\(\bm{LF_3}\))}: it computes the support \(\chi_{j}\) for the class \(\omega_j\) as:
    \begin{equation}
        { \chi  }_{ j } =\min _{ i }{ { \mu }_{ i,j } } 
    \end{equation}
    \item \textbf{Mean rule (\(\bm{LF_4}\))}: it computes the support \(\chi_{j}\) for the class \(\omega_j\) as:
    \begin{equation}
        { \chi  }_{ j } =\frac { 1 }{ L } \sum _{ i=1 }^{ L }{ { \mu }_{ i,j } } 
    \end{equation}
\end{enumerate}

We also investigated others two rules computing the class supports comparing the entire \(DP^{p}\) feature space with the decision templates (DTs) of each class. 
DTs-based methods have been found to be among the best combination techniques and show stable performance over a range of experimental settings~\cite{kuncheva2001decision}.
They are:
\begin{enumerate}
    \item \textbf{Decision Templates, DTs (\(\bm{LF_5}\))}:
    its use was proposed in \cite{kuncheva2001decision} and consists of calculating \(C\) DTs, one per class, that capture the pattern of each. 
    The decision template \(DT_{i}\) for class \(\omega_i\) is the centroid of class \(\omega_i\) in the training \(L\times C\) feature space \(DP^{p}\) and it is calculated as follows: 
    \begin{equation}
        { DT }_{ i }=\frac { 1 }{ { N }_{ i } } \sum _{p=1 }^{N_{i}}{ DP^p  } ,
    \end{equation}
    where \(N_{i}\) is the number of patients belonging to the class \(\omega_i\).
    
    Finally, the \(p\)-th patient's support degree \(\chi_{i}\) for the class \(\omega_i\) is computed by measuring the similarity between the current \(DP^{p} \) and \(DT_{i}\):
    \begin{equation}
        { \chi  }_{ i } =1-\frac { 1 }{ L\cdot C } \sum _{ j=1 }^{ C }{ \sum _{ i=1 }^{ L }{ { \left( { \mu }_{ k,j } -{ dt }_{ k,j }^{ i } \right)  }^{ 2 } }  },
    \end{equation}
    where \({ dt }_{ k,j }^{i}\) is the \(k\), \(j\)-th entry in the \(i\)-th decision template \(DT_{i}\).
    \item \textbf{Dempster-Shafer rule (\(\bm{LF_6}\))}: it is still based on the use of DTs. 
    The \(p\)-th patient's proximity \(\Phi^{p}\) between the output of the \(i\)-th classifier \(D_i^p\) and \(DT_{j}^{i}\) is defined as \cite{kuncheva2004combining}:
    \begin{equation}
        { \Phi  }_{ j,i }^{p} =\frac { { \left( 1+{ \left\| { DT }_{ j }^{ i }-{ D }_{ i }^{p}  \right\|  }^{ 2 } \right)  }^{ -1 } }{ \sum _{ k=1 }^{ C }{ { \left( 1+{ \left\| { DT }_{ j }^{ i }-{ D }_{ i }^{p}  \right\|  }^{ 2 } \right)  }^{ -1 } }  },
    \end{equation}
    where \(DT_{j}^{i}\) denotes the \(i\)-th row of decision template for the class \(\omega_{j}\), \(D_i^p\) denotes the output of the \(i\)-th classifier on the \(p\)-th patient and \(\left\| \cdot  \right\| \) is any matrix norm.
    Then, the final support degree for the \(j\)-th class is:
    \begin{equation}
        { \chi  }_{ j } =K\prod _{ i=1 }^{ L }{ \frac { { \Phi  }_{ j,i }^{p} \prod _{ k\neq j }^{  }{ \left( 1-{ \Phi  }_{ k,i }^{p}  \right)  }  }{ 1-{ \Phi  }_{ j,i }^{p} \left[ 1-\prod _{ k\neq j }^{  }{ \left( 1-{ \Phi  }_{ k,i }^{p}  \right)  }  \right]  }  } 
    \end{equation}
    where \(K\) is a scaling factor.
\end{enumerate}

For the sake of completeness, we also investigated other two rules working with different paradigms.

On the one side, we use the \emph{majority voting rule} (\(LF_7\)) that works with crisp label outputs of each modality by assigning the patient $p$ the class label $\omega_s$ that is most represented among those returned by the $L$ unimodal classifiers. 
Formally: 
\begin{equation}
    s = \argmax _{ j }{ \sum _{ i=1 }^{ L }{ { \mu  }_{ i,j }^{ crisp } }  }, \quad \text{for} \,\,\,  j = 1, \dots, C
\end{equation}
where 
\begin{equation}
    { \mu  }_{ i,j }^{ crisp }=
    \begin{cases}
    1, & \text{if }j=\argmax_{ w }{ { \mu  }_{ i,w } }  \\ 
    \hfill 0, & \text{Otherwise} 
    \end{cases}
    ,\quad \forall \,\,\, i=1,\dots,L \,\, \wedge \,\, j=1,\dots ,C
\end{equation}
On the other side, we also applied the \emph{confidence rule} (\(LF_8\)), which assigns patient \(p\) the class label \(\omega_{s}\):
\begin{equation}
    s = \argmax_{ j }{  DP^p_q }, \quad \text{for} \,\,\, j = 1, \dots, C
\end{equation}
which corresponds to the \(q\)-th unimodal classifier output with the largest degree of support:
\begin{equation}
    q = \argmax_{ i }{ \left( \max { DP^p_i }  \right)  }, \quad \text{for} \,\,\,  i = 1, \dots, L
\end{equation}
where $DP^p_i$ denotes the $i$-th modality whose classifier output is represented by row the $i$-th of $DP^p$. 

\begin{table}[]
\centering
\resizebox{\textwidth}{!}{
\begin{tabular}{lllllllll}
\toprule
& \textbf{Product} 
& \textbf{Max.} 
& \textbf{Min.} 
& \textbf{Mean} 
& \textbf{\makecell{ Decision \\ Template }} 
& \textbf{\makecell{ Dempster \\ Shafer }} 
& \textbf{\makecell{ Majority \\ Vote }} 
& \textbf{Confidence} \\
\midrule
\textbf{\makecell{Features \\ Mean}} 
& \(A_1+LF_1\)     
& \(A_1+LF_2\) 
& \(A_1+LF_3\) 
& \(A_1+LF_4\)  
& \(A_1+LF_5\)               
& \(A_1+LF_6\)             
& \(A_1+LF_7\)           
& \(A_1+LF_8\) \\
\midrule
\textbf{\makecell{Score \\ Mean}}    
& \(A_2+LF_1\)     
& \(A_2+LF_2\) 
& \(A_2+LF_3\) 
& \(A_2+LF_4\)  
& \(A_2+LF_5\)               
& \(A_2+LF_6\)             
& \(A_2+LF_7\)           
& \(A_2+LF_8\)  \\
\bottomrule
\end{tabular}
}
\caption{Summary of the 16 rule combinations performed in this work.}
\label{tb:exps}
\end{table}

\section{Experimental setup}
\label{sc:expresults}
Here we introduce the experimental setup adopted, presenting in section~\ref{subsc:exps} the set of experiments carried out, and in section~\ref{subsc:eval} the validation adapted as well as the evaluation metrics used. 

\subsection{Set of experiments}
\label{subsc:exps}
The first set of experiments consists of evaluating the late fusion paradigm through all the different combinations of the fusion and aggregation rules, \(LF_x\) and \(A_y\), respectively, for a total of 16 combinations since \(x\in\left\{ 1,\dots ,8 \right\}\) and \(y\in\left\{ 1,2 \right\}\), which are summarised in~\autoref{tb:exps}. 
Furthermore, these 16 experiments were performed for all combinations of modalities, i.e. \(\emph{Pathomics} + \emph{Radiomics} + \emph{Semantic}\) (\(P+R+S\)), \(\emph{Pathomics}+\emph{Semantic}\) (\(P+S\)), \(\emph{Radiomics}+\emph{Semantic}\) (\(R+S\)), and, finally, \(\emph{Pathomics}+\emph{Radiomics}\) (\(P+R\)), for a total of 64 experiments.

Then, we compared the late fusion approach with an early fusion framework. 
Concerning this last approach, in this work we considered two early fusion rules: a simple approach in which the different modalities are concatenated without any processing on the feature space, and a concatenation given by the application of the Kronecker product, as presented in~\cite{chen2020pathomic}.
Indeed, the latter rule was chosen to bring out a correlation of the different modalities in the various combinations of them.
For the sake of consistency of samples, in the early fusion paradigm we only applied the A1 aggregation rule, i.e. samples' patient-wise aggregation is performed by averaging each component of the feature vectors belonging to the same patient.

In all the experiments, we used the same learning paradigm in the \emph{classifier} blocks of~\autoref{fig:framework}, in which is a Random Forest~\cite{ho1995random} with entropy as a function to measure the quality of a split, whilst, for all the other parameters, we used the default values provided by the Scikit-learn framework~\cite{sklearn_api}, without any fine-tuning.
Indeed, it was empirically observed in~\cite{arcuri2013parameter} that in many cases the use of tuned parameters cannot significantly outperform the default values of a classifier suggested in the literature, as also confirmed in other works~\cite{zhang2017up,d2019tackling,soda2021aiforcovid}.

Finally, we compared the discrimination strength of hand-crafted features against deep features in the late fusion framework. 
For this comparison, the 64 experiments described above were also performed using deep features extracted from both pathomics and radiomics modalities using the ResNet-18~\cite{he2016deep} and GoogLeNet~\cite{szegedy2015going} networks respectively, pre-trained on ImageNet dataset. 
The choice of using ImageNet as a pre-training tool is motivated by the fact that this dataset provides enough rich image detail of different objects and targets, and therefore we believe that these pre-trained network feature extraction capabilities can be transferred to both pathomics and radiomics tasks.
For each patient we trained the CNNs with a transfer learning process performed with all samples from the other patients for 20 epochs, as suggested by our previous work~\cite{liu2021exploring}.
Furthermore, given the reduced amount of training samples, during the training we froze the weights for all the layers except the ones of the new final fully connected layer.
Straightforwardly in this last layer we removed the original 1000 neurons, which are replaced by two softmax neurons with random weights. These experiments were performed using the PyTorch framework~\cite{paszke2017automatic}.

\subsection{Evaluation methods}
\label{subsc:eval}
We tested all the proposed approaches with a \emph{leave-one-patient-out} (LOPO) cross-validation paradigm so that we performed a number of runs equal to the number of patients.
Therefore, in each run the test set consisted of all samples belonging to one patient, whereas all the others were allocated to the training set.

The patient-wise performances were computed averaging the Area under the ROC curve (AUC)    of each run, where, as a reminder, the positive and negative classes correspond to adaptive and non-adaptive patients, respectively.
It is worth recalling that AUC is a figure of merit widely adopted in the medical community to characterise the performance of a prediction model.
Furthermore, to compare the results we also applied some statistical tests that will be introduced hereinafter, and formally presented in Appendix A.

\section{Results and discussion}
\label{sc:results}

\begin{table}[]
\centering
\begin{tabular}{llll}
\toprule
\multirow{2}{*}{\textbf{\makecell{Aggregation \\ Rules}}} & \multicolumn{3}{c}{\textbf{Modalities}} \\
\cline{2-4}
& \textbf{Pathomics} & \textbf{Radiomics} & \textbf{Semantic}         \\
\midrule
\textbf{-}  & -  & -  & .705    \\
\textbf{A1}  & .686  & \textbf{.870}  & - \\
\textbf{A2}  & .711 & .731 & -   \\
\bottomrule
\end{tabular}
\caption{Results for the unimodal flows in terms of AUC.}
\label{tab:unimodal-res}
\end{table}

\begin{table}[]
    \centering
\begin{tabular}{llcccc}
\toprule
\multirow{2}{*}{\textbf{\makecell{Rules \\ Combination}}} & \multicolumn{4}{c}{\textbf{Modalities Combinations}}      \\
\cline{2-5}
                                   & \textbf{P+R} & \textbf{R+S} & \textbf{P+S} & \textbf{P+R+S} 
\\ \midrule
\(\bm{A_1+LF_1}\)
& \cellcolor{white}.853        & \cellcolor{white}.888        & \cellcolor{white}.752        & \cellcolor{white}.866            \\
\(\bm{A_1+LF_2}\)                     & \cellcolor{white}\textbf{.909}        & \cellcolor{white}.837        & \cellcolor{white}.756        & \cellcolor{white}.860            \\
\(\bm{A_1+LF_3}\)                   & \cellcolor{white}.812        & \cellcolor{white}.864        & \cellcolor{white}.740        & \cellcolor{white}.824            \\
\(\bm{A_1+LF_4}\)                       & \cellcolor{white}.903        & \cellcolor{white}.893        & \cellcolor{white}.764        & \cellcolor{white}.907            \\
\(\bm{A_1+LF_5}\)                    & \cellcolor{white}.888        & \cellcolor{white}\textbf{.909}        & \cellcolor{white}.748        & \cellcolor{white}.905            \\
\(\bm{A_1+LF_6}\)                       & \cellcolor{white}.901        & \cellcolor{white}.781        & \cellcolor{white}.731        & \cellcolor{white}\textbf{.909}            \\
\(\bm{A_1+LF_7}\)                       & \cellcolor{white}.903        & \cellcolor{white}.893        & \cellcolor{white}.764        & \cellcolor{white}.907            \\
\(\bm{A_1+LF_8}\)                     & \cellcolor{white}.853        & \cellcolor{white}.876        & \cellcolor{white}.748        & \cellcolor{white}.857            \\
\(\bm{A_2+LF_1}\)                      & \cellcolor{white}.756        & \cellcolor{white}.752        & \cellcolor{white}.756        & \cellcolor{white}.793            \\
\(\bm{A_2+LF_2}\)                      & \cellcolor{white}.760        & \cellcolor{white}.688        & \cellcolor{white}.684        & \cellcolor{white}.709            \\
\(\bm{A_2+LF_3}\)                    & \cellcolor{white}.715        & \cellcolor{white}.754        & \cellcolor{white}.760        & \cellcolor{white}.777            \\
\(\bm{A_2+LF_4}\)                    & \cellcolor{white}.793        & \cellcolor{white}.752        & \cellcolor{white}.748        & \cellcolor{white}.798            \\
\(\bm{A_2+LF_5}\)                   & \cellcolor{white}.789        & \cellcolor{white}.756        & \cellcolor{white}.748        & \cellcolor{white}.810            \\
\(\bm{A_2+LF_6}\)                     & \cellcolor{white}.769        & \cellcolor{white}.740        & \cellcolor{white}.731        & \cellcolor{white}.773            \\
\(\bm{A_2+LF_7}\)                      & \cellcolor{white}.793        & \cellcolor{white}.752        & \cellcolor{white}.748        & \cellcolor{white}.798            \\
\(\bm{A_2+LF_8}\)                     & \cellcolor{white}.715        & \cellcolor{white}.740        & \cellcolor{white}.740        & \cellcolor{white}.764   \\
\bottomrule
\end{tabular}

\caption{Overall results for the 64 experiments performed in terms of AUC, where \emph{P}, \emph{R} and \emph{S} stand for pathomics, radiomics and semantic, respectively.}
    \label{tab:overall-multimodalres}
\end{table}

This section presents and analyses the results in several directions, starting from the raw outputs reported in \autoref{tab:unimodal-res} and \autoref{tab:overall-multimodalres}. 
The former reports the scores attained by each single modality, eventually  using one aggregation rule to combine the features describing the samples.
The radiomics modality with the use of the feature mean as patient aggregation rule is the best unimodal flow with an AUC equal to \(0.870\).
The latter shows the performance attained by the pairwise fusion approaches and by the trimodal combination, for all the 16 fusion rules.
With an AUC equal to \(0.909\), the best results are achieved by the multimodal triplet \(P+R+S\), and the pairwise combinations \(R+S\) and \(P+R\), with the following fusion rules respectively: \(A_1+LF_6\), \(A_1+LF_5\) and \(A_1+LF_2\). 
Hence, all of them are given by the use of the feature mean as patient aggregation rule at feature level followed by the Dempster-Shafer, Decision Template and Maximum as fusion rule, respectively.

To discuss these results, the rest of this section deepens the results in three directions. 
First, we present the results provided by the late fusion approaches introduced in  section~\ref{subsc:fusion-rule} and schematically depicted in \autoref{fig:framework} (section~\ref{subsc:late}).
Second, in section~\ref{subsc:hand-deep} we compare late fusion and early fusion approaches.
Third, in section~\ref{subsc:hand-deep} we show a comparison of hand-crafted and deep features.

\begin{figure}
    \centering
    \includegraphics[scale=0.4]{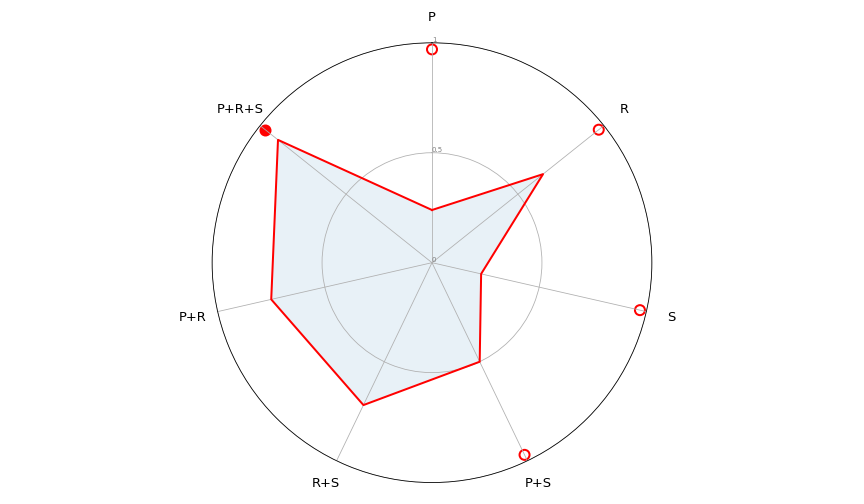}
    \caption{Radar chart showing the performance in terms of AUC of the unimodal and multimodal approaches, where P stands for Pathomics, R for Radiomics, and S for Semantic.
    The filled circle represents the flow with the highest rank, whilst blank circles represent unimodal or multimodal approaches with statistically different performances from the best approach according to Friedman test with the Iman-Davenport amendment followed by the pairwise Bonferroni-Dunn post-hoc test (\(p<0.1\)).}
    \label{fig:bonferroni_latefusion}
\end{figure}

\begin{figure}
    \centering
    \includegraphics[scale=0.4]{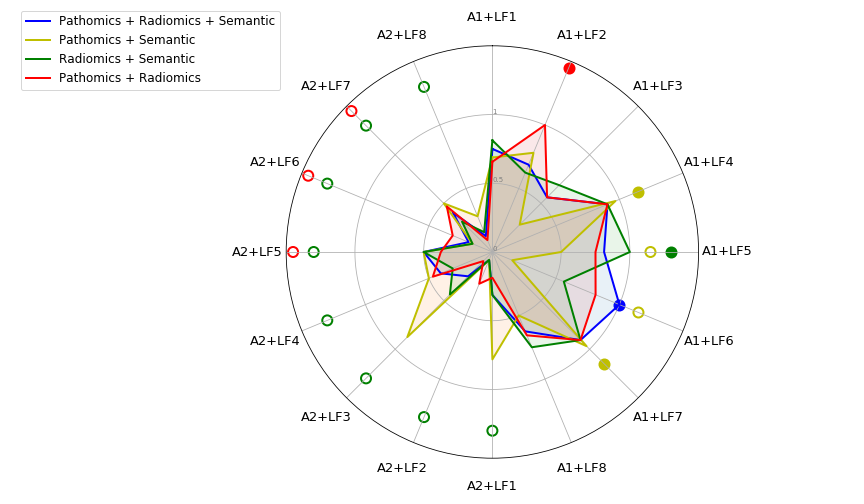}
    \caption{Radar chart showing the performance in terms of AUC of the fusion rules given by the combination of \(LF_x\) and \(A_y\) (\autoref{fig:framework}) and varying the modalities combination.
    Filled circles represent the fusion rule combination with the highest rank, whilst blank circles represent models with statistically different performances from the best model according to the Wilcoxon signed-rank test (\(p < 0.1\)).}
    \label{fig:bestrule_radarplot}
\end{figure}

\subsection{Late fusion results}
\label{subsc:late}
The contribution of this experiment is three-fold, so it permits us to answer the following three questions: 
\begin{enumerate}[noitemsep]
    \item What is the best combination of modes? 
    \item Within the multimodal combination, which is the unimodal mode contributing more to the best performance?
    \item What is the best fusion rule? 
\end{enumerate}
Let us now explore each of these questions.
In all the cases we will introduce figures that offer a high-level synthesis of the huge amount of results provided by all the experiments.

\paragraph{\textbf{Best multimodal combination}}
\autoref{fig:bonferroni_latefusion} shows a radar chart plotting the performances in terms of AUC of the various unimodal and multimodal approaches. 
As mentioned above and summarised in~\autoref{tb:exps} we have a total of 16 different rules, given by the different combinations of the aggregation rules (\(A_y\)) and the late fusion rules (\(LF_x\)).
So for each of these  rule combinations we rank each approach so that the one with the highest performance receives a score of 7, whilst the one with the lowest performance gets a score of 1. 
At the end of this iterative process the rank of each stream is given by the sum of the ranks received for each of the 16 experiments mentioned above. We then normalise for the maximum rank achievable.
In the figure, we adopt a  filled circle to mark the flow with the highest rank, while the blank circles denote those approaches with a lower rank, whose performances are statistically different from the best one according to the Friedman test with the Iman-Davenport amendment followed by the Bonferroni-Dunn pairwise post-hoc test (\(p<0.1\)).
Furthermore, we do not report any circle when the rank of  a flow is lower than the best one and the corresponding   performance are not statistically different.

Under these premises, in this figure the lengths of the spokes show how the multimodal approaches generally perform better than the unimodal ones, as the latter always differ from the best combination in a statistically significant way.
Furthermore, as you can see from the filled circle, the trimodal combination given by \(P + R + S\) is the best approach and it significantly differs from all unimodal approaches (i.e. \(P\), \(R\), \(S\))  \(P + S\). 
Although in a different clinical context, similar considerations about the inherent superiority of multimodal approaches over the unimodal flows were obtained from~\cite{braman2021deep, grossmann2017defining}. 
Indeed in \cite{braman2021deep} and \cite{grossmann2017defining}, the overall survival analyses  performed for glioma and lung cancer, respectively, show how the multimodal combination significantly outperforms the best performing unimodal flows.
Furthermore, as in our work, in both studies the triplet multimodal combination emerged as the best approach.

\paragraph{\textbf{Most informative unimodal approach}}
Let us now rank the unimodal approaches in order to understand which flow is the most informative in terms of AUC. 
The ranks are computed as described before.
Indeed, given a fusion rule, the multimodal approaches are ranked so that the flow with the best performance gets rank 4, as we have four different multimodal approaches.
Then, all the unimodal approaches that make up the multimodal flow get the same rank as the multimodal one.
So, for instance, if the case \(P+R+S\) multimodal flow got rank \(3\), the unimodal approaches of pathomics, radiomics and semantic all get the rank \(3\). 
This operation is repeated for all 16 different rule combinations and the final ranks are updated by accumulation and, finally, normalised for the maximum rank achievable. 

Hence, with a rank equal to  \(0.667\), the radiomics approach is the most informative unimodal flow, outperforming both the pathomics and semantic flows which got a rank equal to \(0.549\) and \(0.535\), respectively.
Similar considerations were obtained from~\cite{braman2021deep} where, although in  a different clinical context, the overall survival analysis of glioma dealing with the radiomics unimodal flow emerges as the most informative modality in terms of Cox Loss.

\paragraph{\textbf{Best fusion rule}}
\autoref{fig:bestrule_radarplot} shows a radar chart plotting the  performance  in  terms  of AUC of the various fusion rules varying the way we combine the modalities.
Let us remember that, as mentioned above and summarised in \autoref{tb:exps}, the rules represented in the figure are the combination of the aggregation and late fusion rules, \(A_y\) and \(LF_x\), respectively.
For each multimodal combination, we ranked each fusion rule in terms of AUC so that the worst rules receives rank 1, whilst the best receives rank 16.
Filled circles in the figure represent the fusion rule combination with the highest rank, whilst blank circles represent models with statistically different performances from the best fusion rule according to the Wilcoxon signed-rank test (\(p < 0.1\)).
Note that here we  used such test rather than Friedman's method since, for each late fusion rule, we compared the four different multimodal combinations, a number limiting the application of Friedman's test.
Furthermore, we do not report any circle when the rank of  a flow is lower than the best one and the corresponding   performance are not statistically different.

The chart shows that the patient aggregation rule named as score mean aggregation (\(A_2\)) performs generally worst than the feature mean rule (\(A_1\)). 
On the other hand, if we focus on the late fusion rules, we observe that they perform almost equally well on all multimodal combinations.

Moreover, the best combinations of rules depend on the modality combination considered. 
For the \(P+R+S\) combination, the best rule combination is denoted as \(A_1 + LF_6\), which is therefore given by the use of the feature mean as patient aggregation rule at feature level followed by the Dempster Shafer as fusion rule working on
the outputs of each classifier.
For the \(P+S\) combination, the best rules are denoted as \(A_1 + LF_4\) and \(A_1 + LF_7\), which are given by the combination of the feature mean as patient aggregation rule and, respectively, the mean and majority vote as fusion rule.
For the \(R+S\) combination, the best rule combination is denoted as \(A_1 + LF_5\), which is therefore given by the use of the feature mean followed by the Decision Template as fusion rule.
For the \(P+R\) combination, the best rule combination is denoted as \(A_1 + LF_2\), which is therefore given by the use of the feature mean followed by the maximum as fusion rule.
Globally, the best fusion rules are \(A_1 + LF_4\) and \(A_1 + LF_7\), since on average they ranked highest across all modality combinations.
It suggests that these rules are well adapted to different situations,    generalising successfully across different types of datasets, which, in turn, are characterised by different data types, sizes and dimensionalities.

\begin{table}[]
\centering
\begin{tabular}{lcc}
\toprule
\textbf{Multimodal Combination} & \textbf{Simple Concatenation} & \textbf{Kronecker} \\
\midrule
Pathomics + Radiomics + Semantic & \cellcolor{white} 5-2-1 & \cellcolor[HTML]{C0C0C0} 8-0-0\\
Pathomics + Semantic  & \cellcolor[HTML]{C0C0C0} 8-0-0 & \cellcolor[HTML]{C0C0C0} 8-0-0 \\
Radiomics + Semantic &  \cellcolor{white} 4-1-3 &  \cellcolor[HTML]{C0C0C0} 7-0-1 \\
Pathomics + Radiomics  &  \cellcolor{white} 5-0-3 & \cellcolor[HTML]{C0C0C0} 7-1-0 \\
\bottomrule
\end{tabular}
\caption{Exhaustive comparison of late and early multimodal learning for the various modality combinations in terms of AUC. 
Each cell shows the amount of win–tie–loss of a combination of the corresponding multimodal combination handled in the late fusion paradigm.
For each modalities combination, the  cells  are highlighted  in  grey  when the late fusion approaches are significantly better than the early fusion rules according to the one-tailed sign test (\(p<0.05\)).}
\label{tb:earlyfusion}
\end{table}

\subsection{Late vs Early fusion}
\label{subsc:latevsearly}
\autoref{tb:earlyfusion} shows an exhaustive comparison of late and early multimodal learning for the various modality combinations in terms of AUC.
As we already mentioned, with the early fusion paradigm we only tested the \(A_1\) aggregation rule for the sake of consistency of samples.
Each cell reports the amount of win–tie–loss of the corresponding multimodal flow handled in the late fusion paradigm. Since early fusion only handles the \(A_1\) aggregation rule, the comparison was restricted to only the 8 late fusion rules that use this method of aggregation.
Given the number of patient data, we statistically validated this comparison with the sign test, a simple but powerful statistical test.
In \autoref{tb:earlyfusion}, for each modalities combination, the grey cells highlight when the late fusion approaches are significantly better than early ones according to the one-tailed sign test (\(p <0.05\)).
The table shows that the late fusion paradigm almost always outperforms the early fusion paradigm for the performance metric considered.
This suggests us that there is a certain difficulty in the data-level fusion process when the data are significantly uncorrelated, and have such a different nature and dimensionality, as in the medical task we are dealing with in this work.

\begin{table}[]
\centering
\renewcommand{\arraystretch}{1.2}
\begin{tabular}{c|lcccc}
\toprule
\multicolumn{1}{l}{}                            & \multicolumn{1}{l}{} & \multicolumn{4}{c}{\textbf{Deep Features}}        \\
\cmidrule{3-6}
\multicolumn{1}{l}{}                            &                      & \textbf{P+R+S}      & \textbf{P+S}        & \textbf{R+S}        & \textbf{P+R}        \\ 
\multirow{4}{*}[2ex]{\rotcell[c]{\textbf{Hand-crafted}\\\textbf{Features}}} & \textbf{P+R+S}               & \cellcolor[HTML]{C0C0C0} 16-0-0 & 
\cellcolor[HTML]{C0C0C0} 16-0-0  & 
\cellcolor[HTML]{C0C0C0} 16-0-0  & 
\cellcolor[HTML]{C0C0C0} 16-0-0 \\
& \textbf{P+S}  & 
\cellcolor{white}        11-0-5  & 
\cellcolor[HTML]{C0C0C0} 15-1-0  &
\cellcolor{white}        10-3-3  & 
\cellcolor[HTML]{C0C0C0} 16-0-0 \\
& \textbf{R+S} & 
\cellcolor[HTML]{C0C0C0} 16-0-0  &
\cellcolor[HTML]{C0C0C0} 16-0-0  & 
\cellcolor[HTML]{C0C0C0} 14-2-0  &
\cellcolor[HTML]{C0C0C0} 16-0-0 \\
& \textbf{P+R} & 
\cellcolor[HTML]{C0C0C0} 16-0-0  & 
\cellcolor[HTML]{C0C0C0} 16-0-0  & 
\cellcolor[HTML]{C0C0C0} 13-1-2 
& \cellcolor[HTML]{C0C0C0} 16-0-0  \\
\bottomrule
\end{tabular}
\caption{Exhaustive comparison between the performance of modality combinations expressed in terms of AUC, where P stands for Pathomics, R for Radiomics, and S for Semantic.
Each cell shows the amount of win–tie–loss of a combination in a row compared with a combination in a column, respectively performed using hand-crafted and deep features.
The cells highlighted in grey represent the modality combination significantly better than another according to the one-tailed sign test (\(p<0.05\)).
}
\label{tb:exhandvsdeep}
\end{table}

\subsection{Hand-crafted vs Deep Features}
\label{subsc:hand-deep}

This experiment compares hand-crafted and deep features in the same multimodal late fusion framework (\autoref{fig:example-images}).
Hence, for the two descriptor groups we have the same number of experiments, i.e. 64 as discussed in section~\ref{subsc:exps}.

We summarised the comparison in \autoref{tb:exhandvsdeep}, which offers an exhaustive comparison between the modality combinations in terms of AUC.
Each cell shows the amount of win–tie–loss of a pair of modality combinations indexed by row and column, respectively performed using hand-crafted and deep features.
For instance, the second cell in row 1, with indexes \(\left(1,2\right)\), counts the wins, ties and losses obtained by the modalities triplet \(P+R+S\) performed with hand-crafted features against the modalities pair \(P+S\) performed with deep features.
Since for each combination we have a total of 16 different combinations of the fusion and aggregation rules, \(LF_x\) and \(A_y\) respectively, the total amount of scores for each cell is equal to 16.

Again, we statistically validated this comparison with the sign test.
In \autoref{tb:exhandvsdeep} the grey cells represent the modality combination significantly better than another according to the one-tailed sign test (\(p <0.05\)).
From the table we can see how the hand-crafted features perform better than the deep features. 
The reason for this result can be found in the low dimensionality of the dataset, as could be expected. 
Indeed, this limits the ability of the deep neural networks to fully express their power of abstraction, generalisation and discrimination.

\section{Conclusion}
\label{sc:conclusion}

Nowadays, lung cancer is worldwide recognised as the most common type of cancer and one of the most frequent causes of tumour death despite the recent increase in the number of treatment options. 
The current clinical decision-making process relies on multiple data sources to improve the prognostic outcomes of radiotherapy treatment, such as radiology-based data, digital pathology slides, genome profiling and clinical data.
Despite the importance of these data sources, to the best of our knowledge only one work to date has combined them together into a single deep learning framework fed with a large amount of data collected from public repositories~\cite{braman2021deep}.
Although deep learning networks offer a high generalisation ability, they require a large amount of data that are not always available in every clinical context.
Hence, in this work, we have presented a multimodal late fusion framework combing radiomics, pathomics and clinical data to predict radiation therapy treatment outcomes for NSCLC patients.
We fed the proposed framework with hand-crafted features extracted from the aforementioned data sources.
Here, we explored the combinations of eight different late fusion rules (i.e. product, maximum, minimum, mean, decision template, Dempster-Shafer, majority voting, and confidence rule) with two samples’ patient-wise aggregation rules (i.e. feature mean and score mean) implemented to have a single classification per patient and consistent sample fusion, for a total a 64 experiments.
The contribution of these experiments is three-fold, as they allow us to find the best combination of modalities, the most informative unimodal flow and, finally, the best fusion rule. 
From these experiments emerges that the trimodal combination given by pathomics, radiomics and semantic data (i.e. \(P + R + S\)) is the best approach and it significantly differs from all unimodal approaches and the pairwise combination \(P + S\).
Furthermore, the radiomics approach is the most informative unimodal flow, outperforming both the pathomics and semantic flows.
Finally, this set of experiments shows that the best fusion rules are \(A_1 + LF_4\) and \(A_1 + LF_7\), which are given by the combination of the feature mean as patient aggregation rule and, respectively, the mean and majority vote as fusion rule, since on average they ranked highest across all modality combinations. This suggests us that these rules are well suited to different situations, successfully generalising across different types of modalities.

As another contribution of this work, we compared the late fusion framework with an early one.
Concerning the latter, we explored two early fusion rules: a simple concatenation in which the different modalities are concatenated without any processing on the feature space, and a concatenation processed by the application of the Kronecker product.
The preliminary results show that the early approach leads to a deterioration in performance due to their lack of capacity to handle highly heterogeneous data such as in the medical task dealt with in this work.

As a final contribution, we fed the multimodal late fusion scheme also with deep features extracted from pre-trained deep neural networks fine-tuned with the stand-alone modalities. 
As could be expected due to the limited size of the patients’ cohort, the results show as the deep features lead to a general performance deterioration.
Nevertheless, also in this case the combination of the three modalities leads to an improvement in performance.

Finally, turning to the clinical side, the take-home message emerging from this work is that the multimodal learning framework leads to a significant improvement of a learning system in terms of performance.
Indeed, in this work the simultaneous fusion of the three modalities given by \(P + R + S\) is the best approach and it significantly differs from all the models fed with the stand-alone data flows. 
Although in a different clinical context, similar considerations about the inherent superiority of multimodal approaches over the unimodal ones were obtained from \cite{braman2021deep, grossmann2017defining}.

\section*{Conflict of interest statement}
All authors declare that they have no known competing financial interests or personal relationships that could have appeared to influence (bias) the work reported in this paper.

\section*{Author contributions}
\textbf{Matteo Tortora}: Conceptualization, Methodology, Software, Validation, Formal analysis, Investigation, Resources, Data Curation, Writing - Original Draft, Writing - Review \& Editing, Visualization.
\textbf{Ermanno Cordelli}: Conceptualization, Methodology, Software, Validation,  Investigation, Resources, Data Curation, Writing - Review \& Editing.
\textbf{Rosa Sicilia}: Conceptualization, Data Curation, Writing - Review \& Editing.
\textbf{Lorenzo Nibid}: Data Curation, Writing - Review \& Editing.
\textbf{Edy Ippolito}: Data Curation, Writing - Review \& Editing.
\textbf{Giuseppe Perrone}: Conceptualization, Resources, Data Curation, Writing - Review \& Editing.
\textbf{Sara Ramella}: Conceptualization, Resources, Data Curation, Writing - Review \& Editing,  Project administration, Funding acquisition.
\textbf{Paolo Soda}: Conceptualization, Methodology, Validation, Resources, Formal analysis, Investigation, Writing - Original Draft, Writing - Review \& Editing, Visualization, Supervision, Project administration, Funding acquisition.

\section*{Acknowledgements}
This work was partially founded by Università Campus Bio-Medico di Roma  under the programme ``University Strategic Projects'', within the project ``a CoLlAborative multi-sources Radiopathomics approach for personalized Oncology in non-small cell lung cancer (CLARO)'', by PON ``Ricerca e Innovazione 2014- 2020,  Azioni IV.4 –  Dottorati e contratti di ricerca su tematiche dell’innovazione'' and by Regione Lazio under the program ``PO FSE 2014-2020  Azione Cardine 21''.

\input{appendix}

\balance
\bibliographystyle{unsrtnat}
\bibliography{references.bib}

\end{document}

%% file: appendix.tex
\appendix
\section{Statistical tests}
\label{app:statisticaltest}
The following sections describe the statistical tests we used in this work to validate and compare our approaches.

\subsection*{\textbf{Friedman test with Iman and Davenport amendment}}
The Friedman test with amendment by Iman and Davenport is a non-parametric statistical test used to compare multiple models.
For each of the 16 different combinations of fusion rules (\(A_y+LF_x\)), each flow is ranked  so that the best receives rank 1, whilst the worst receives rank 7. 
Tied ranks are shared equally as explained above.
The test statistic with the amendment proposed by Iman and Davenport is the following:
\begin{equation}
    { F }_{ F }=\frac { \left( N-1 \right) { \chi  }_{ F }^{ 2 } }{ N\left( M-1 \right) -{ \chi  }_{ F }^{ 2 } } 
\end{equation}
which follows the \(F\)-distribution with \(\left( M-1 \right) \) and \(\left( M-1 \right) \left( N-1 \right) \) degrees of freedom and where:
\begin{equation}
    { \chi  }_{ F }^{ 2 }=\frac { 12N }{ M\left( M+1 \right)  } \left( \sum _{ j=1 }^{ M }{ { R }_{ j }^{ 2 } } -\frac { M{ \left( M+1 \right)  }^{ 2 } }{ 4 }  \right) 
\end{equation}
where \({ R }_{ j }=\frac { 1 }{ N } \sum _{ i=1 }^{ N }{ { r }_{ i }^{ j } } \) is the average rank of the \(j\)-th flow and \(r_i^j\) is the rank of the \(j\)-th flow when considering the \(i\)-th fusion rule, where \(i=1,\dots ,N\) and \(j=1,\dots ,M\) (\(N=16, M=7\)).
Once the \(F\)-statistic has been computed, we can carry out the test comparing it with the critical value for the chosen level of significance.  If it is greater than this value we can reject the null hypothesis \(H_0\) and accept that there is a difference between the flows.

\paragraph{\textbf{Bonferroni-Dunn post-hoc test}}
If \(H_0\) is rejected, Bonferroni-Dunn post-hoc test is applied to find exactly where the differences are:
\begin{equation}
    z=\frac { { R }_{ 1 }-{ R }_{ j } }{ \sqrt { \frac { M\left( M+1 \right)  }{ 6N }  }  } 
\end{equation}
where \({ R }_{ 1 }\) is the average rank of the best flow and \({ R }_{ j }\) is the average rank of \(j\)-th flow.
Two flows are statistically different if the obtained \(p\)-value from this \(z\)-value is smaller than \(\frac { \alpha }{ M-1 } \), where \(\alpha\) is the desired level of significance.

\subsection*{\textbf{Wilcoxon signed rank test}}
The Wilcoxon signed-rank test is a non-parametric statistical test that tests if two models are statistically different. 
Given the error estimates of two models for the \(N\) folds of the LOPO validation paradigm, the test computes the differences of these errors \(d_i\).
Then it ranks the absolute values of the differences \(\left| d_i \right| \) so that the smallest value receives rank \(1\), whilst the largest one receives rank \(N\). If there is a tie, all the ranks are shared so that the total sum stays \(1 + 2 +\dots + N\).
Subsequently, it splits the ranks into positive and negative according to the sign of \(d_i\) and calculates the following amounts:
\begin{equation}
    { R }_{ + }=\sum _{ { d }_{ i }>0 }^{  }{ { r }_{ i } } +\frac { 1 }{ 2 } \sum _{ { d }_{ i }=0 }^{  }{ { r }_{ i } } ,\qquad { R }_{ - }=\sum _{ { d }_{ i }<0 }^{  }{ { r }_{ i } } +\frac { 1 }{ 2 } \sum _{ { d }_{ i }=0 }^{  }{ { r }_{ i } } 
\end{equation}
Finally, the test computes the following statistic:
\begin{equation}
    z=\frac { T-\frac { 1 }{ 4 } N\left( N+1 \right)  }{ \sqrt { \frac { 1 }{ 24 } N\left( N+1 \right) \left( 2N+1 \right)  }  } 
\end{equation}
which is approximately distributed as a normal distribution and where \(T=\min { \left( { R }_{ + },{ R }_{ - } \right)  }\). 
Once  the \(z\)-statistic  has  been  computed,  we can carry out the test comparing it with the critical value for the chosen level of significance. 
If it is greater than this value we can reject the null hypothesis and state that the two methods have statistically different performances.

\subsection*{\textbf{Sign test}}
The sign test is simply performed counting wins, ties and losses, with or without statistical significance, of each method pair.
This test is based on the intuition that if two methods are equivalent, each one will perform better than the other one on approximately \({ N }/{ 2 }\) of the tests.
Hence, following the binomial distribution, we can claim that the first method is significantly better than the second one if its amount of wins is greater than \({ N }/{ 2 }+1.96\sqrt { { N }/{ 2 } } \), at a level of significance of \(0.05\).

\section{Statistical descriptors}
\label{app:haralick}
The following are the statistical measures extracted from the intensity histogram and used in this work to provide a synthetic description of both the LBP and the grey level histogram: 
\begin{itemize}[noitemsep]
    \item \textbf{Mean}:
    \begin{equation*}
        m=\sum _{ k=0 }^{ l-1 }{ { r }_{ k }p\left( { r }_{ k } \right)  } 
    \end{equation*}
    \item \textbf{Standard deviation}:
    \begin{equation*}
        \sigma =\sqrt { \sum _{ k=0 }^{ l-1 }{ { \left( { r }_{ k }-m \right)  }^{ 2 }p\left( { r }_{ k } \right)  }  } 
    \end{equation*}
    \item \textbf{Smoothness}:
    \begin{equation*}
        R=1-\frac { 1 }{ { \left( 1+{ \sigma  }^{ 2 } \right)  } } 
    \end{equation*}
    \item \textbf{Skewness}:
    \begin{equation*}
        \text{skewness}=\sum _{ k=0 }^{ l-1 }{ { \left( { r }_{ k }-m \right)  }^{ 3 }p\left( { r }_{ k } \right)  } 
    \end{equation*}
    \item \textbf{Kurtosis}:
    \begin{equation*}
        \text{kurtosis}=\sum _{ k=0 }^{ l-1 }{ { \left( { r }_{ k }-m \right)  }^{ 4 }p\left( { r }_{ k } \right)  } 
    \end{equation*}
    \item \textbf{Energy}:
    \begin{equation*}
        \text{energy}=\sum _{ k=0 }^{ l-1 }{ { p\left( { r }_{ k } \right)  }^{ 2 } } 
    \end{equation*}
    \item \textbf{Entropy}:
    \begin{equation*}
        \text{entropy}=-\sum _{ k=0 }^{ l-1 }{ { p\left( { r }_{ k } \right)  }\log _{ 2 }{ \left[ p\left( { r }_{ k } \right)  \right]  }  } 
    \end{equation*}
    \item \textbf{Absolute maximum}:
    \begin{equation*}
        \max_{k=0}^{l-1}\left[p\left( { r }_{ k } \right)\right]
    \end{equation*}
    \item \textbf{Maximum value}:
    \begin{equation*}
        \argmax_{k=0}^{l-1}\left[p\left( { r }_{ k } \right)\right]
    \end{equation*}
\end{itemize}
where \(l\) is the number of grey levels in the image \(I\) and \(p\left( { r }_{ k } \right)\) is the number of pixels with a grey level equal to \({ r }_{ k }\).

Given instead a grey-level co-occurrence matrix \(G\) extracted from image \(I\), the following are the formal definition of Haralick features used in this work:
\begin{itemize}[noitemsep]
    \item \textbf{Contrast}: 
    \begin{equation*}
        \sum_{i,j=0}^{l-1} G_{i,j}\left(g_i-g_j\right)^2
    \end{equation*}

    \item \textbf{Dissimilarity}:
    \begin{equation*}
        \sum_{i,j=0}^{l-1}G_{i,j}\left|g_i-g_j\right|
    \end{equation*}
    
    \item \textbf{Homogeneity}:
    \begin{equation*}
        \sum_{i,j=0}^{l-1}\frac{G_{i,j}}{1+(g_i-g_j)^2}
    \end{equation*}
    
    \item \textbf{Angular Second Moment (ASM)}: 
    \begin{equation*}
        \sum_{i,j=0}^{l-1} G_{i,j}^2
    \end{equation*}
    
    \item \textbf{Energy}: 
    \begin{equation*}
        \sqrt{ASM}
    \end{equation*}
    
    \item \textbf{Correlation}: 
    \begin{equation*}
        \sum_{i,j=0}^{l-1} G_{i,j}\left[\frac{(g_i-\mu_i) \
(g_j-\mu_j)}{\sqrt{(\sigma_i^2)(\sigma_j^2)}}\right]
    \end{equation*}
\end{itemize}
where \(G_{i,j}\) denotes the \(i,j\)-th entry of \(G\), \(l\) is the number of grey levels in the image \(I\), \(g_i\) and \(g_j\) denote two grey level values \(\in \left[ 0, 2^l-1 \right]\), and, finally, \(\mu_i\) denotes the mean value of the one-dimensional marginal distributions of \(G\).

%% file: main.bbl
\begin{thebibliography}{45}
\providecommand{\natexlab}[1]{#1}
\providecommand{\url}[1]{\texttt{#1}}
\expandafter\ifx\csname urlstyle\endcsname\relax
  \providecommand{\doi}[1]{doi: #1}\else
  \providecommand{\doi}{doi: \begingroup \urlstyle{rm}\Url}\fi

\bibitem[Bray et~al.(2018)Bray, Ferlay, Soerjomataram, Siegel, Torre, and
  Jemal]{bray2018global}
Freddie Bray, Jacques Ferlay, Isabelle Soerjomataram, Rebecca~L Siegel,
  Lindsey~A Torre, and Ahmedin Jemal.
\newblock Global cancer statistics 2018: Globocan estimates of incidence and
  mortality worldwide for 36 cancers in 185 countries.
\newblock \emph{CA: a cancer journal for clinicians}, 68\penalty0 (6):\penalty0
  394--424, 2018.

\bibitem[Lee et~al.(2017)Lee, Lee, Park, Schiebler, van Beek, Ohno, Seo, and
  Leung]{lee2017radiomics}
Geewon Lee, Ho~Yun Lee, Hyunjin Park, Mark~L Schiebler, Edwin~JR van Beek,
  Yoshiharu Ohno, Joon~Beom Seo, and Ann Leung.
\newblock Radiomics and its emerging role in lung cancer research, imaging
  biomarkers and clinical management: state of the art.
\newblock \emph{European Journal of Radiology}, 86:\penalty0 297--307, 2017.

\bibitem[Lambin et~al.(2012)Lambin, Rios-Velazquez, Leijenaar, Carvalho,
  Van~Stiphout, Granton, Zegers, Gillies, Boellard, Dekker,
  et~al.]{lambin2012radiomics}
Philippe Lambin, Emmanuel Rios-Velazquez, Ralph Leijenaar, Sara Carvalho,
  Ruud~GPM Van~Stiphout, Patrick Granton, Catharina~ML Zegers, Robert Gillies,
  Ronald Boellard, Andr{\'e} Dekker, et~al.
\newblock Radiomics: extracting more information from medical images using
  advanced feature analysis.
\newblock \emph{European Journal of Cancer}, 48\penalty0 (4):\penalty0
  441--446, 2012.

\bibitem[Kumar et~al.(2012)Kumar, Gu, Basu, Berglund, Eschrich, Schabath,
  Forster, Aerts, Dekker, Fenstermacher, et~al.]{kumar2012radiomics}
Virendra Kumar, Yuhua Gu, Satrajit Basu, Anders Berglund, Steven~A Eschrich,
  Matthew~B Schabath, Kenneth Forster, Hugo~JWL Aerts, Andre Dekker, David
  Fenstermacher, et~al.
\newblock Radiomics: the process and the challenges.
\newblock \emph{Magnetic Resonance Imaging}, 30\penalty0 (9):\penalty0
  1234--1248, 2012.

\bibitem[Gupta et~al.(2019)Gupta, Kurc, Sharma, Almeida, and
  Saltz]{gupta2019emergence}
Rajarsi Gupta, Tahsin Kurc, Ashish Sharma, Jonas~S Almeida, and Joel Saltz.
\newblock The emergence of pathomics.
\newblock \emph{Current Pathobiology Reports}, 7\penalty0 (3):\penalty0 73--84,
  2019.

\bibitem[Zhang et~al.(2020)Zhang, He, Guo, Liu, Yang, Zhang, Xie, Mu, Guo, Fu,
  et~al.]{zhang2020novel}
Yiying Zhang, Kan He, Yan Guo, Xiangchun Liu, Qi~Yang, Chunyu Zhang, Yunming
  Xie, Shengnan Mu, Yu~Guo, Yu~Fu, et~al.
\newblock A novel multimodal radiomics model for preoperative prediction of
  lymphovascular invasion in rectal cancer.
\newblock \emph{Frontiers in Oncology}, 10:\penalty0 457, 2020.

\bibitem[Wu et~al.(2021)Wu, Ma, Huang, Ling, and Su]{wu2021deepmmsa}
Yujiao Wu, Jie Ma, Xiaoshui Huang, Sai~Ho Ling, and Steven~Weidong Su.
\newblock Deepmmsa: A novel multimodal deep learning method for non-small cell
  lung cancer survival analysis.
\newblock \emph{arXiv preprint arXiv:2106.06744}, 2021.

\bibitem[Chen et~al.(2020)Chen, Lu, Wang, Williamson, Rodig, Lindeman, and
  Mahmood]{chen2020pathomic}
Richard~J Chen, Ming~Y Lu, Jingwen Wang, Drew~FK Williamson, Scott~J Rodig,
  Neal~I Lindeman, and Faisal Mahmood.
\newblock Pathomic fusion: an integrated framework for fusing histopathology
  and genomic features for cancer diagnosis and prognosis.
\newblock \emph{IEEE Transactions on Medical Imaging}, 2020.

\bibitem[Braman et~al.(2021)Braman, Gordon, Goossens, Willis, Stumpe, and
  Venkataraman]{braman2021deep}
Nathaniel Braman, Jacob~WH Gordon, Emery~T Goossens, Caleb Willis, Martin~C
  Stumpe, and Jagadish Venkataraman.
\newblock Deep orthogonal fusion: Multimodal prognostic biomarker discovery
  integrating radiology, pathology, genomic, and clinical data.
\newblock In \emph{International Conference on Medical Image Computing and
  Computer-Assisted Intervention}, pages 667--677. Springer, 2021.

\bibitem[Huang et~al.(2020)Huang, Pareek, Seyyedi, Banerjee, and
  Lungren]{huang2020fusion}
Shih-Cheng Huang, Anuj Pareek, Saeed Seyyedi, Imon Banerjee, and Matthew~P
  Lungren.
\newblock Fusion of medical imaging and electronic health records using deep
  learning: a systematic review and implementation guidelines.
\newblock \emph{NPJ Digital Medicine}, 3\penalty0 (1):\penalty0 1--9, 2020.

\bibitem[Baltru{\v{s}}aitis et~al.(2018)Baltru{\v{s}}aitis, Ahuja, and
  Morency]{baltruvsaitis2018multimodal}
Tadas Baltru{\v{s}}aitis, Chaitanya Ahuja, and Louis-Philippe Morency.
\newblock Multimodal machine learning: A survey and taxonomy.
\newblock \emph{IEEE Transactions on Pattern Analysis and Machine
  Intelligence}, 41\penalty0 (2):\penalty0 423--443, 2018.

\bibitem[Ramachandram and Taylor(2017)]{ramachandram2017deep}
Dhanesh Ramachandram and Graham~W Taylor.
\newblock Deep multimodal learning: A survey on recent advances and trends.
\newblock \emph{IEEE Signal Processing Magazine}, 34\penalty0 (6):\penalty0
  96--108, 2017.

\bibitem[Kiela et~al.(2018)Kiela, Grave, Joulin, and
  Mikolov]{kiela2018efficient}
Douwe Kiela, Edouard Grave, Armand Joulin, and Tomas Mikolov.
\newblock Efficient large-scale multi-modal classification.
\newblock In \emph{Thirty-Second AAAI Conference on Artificial Intelligence},
  2018.

\bibitem[Cavalcanti et~al.(2016)Cavalcanti, Oliveira, Moura, and
  Carvalho]{cavalcanti2016combining}
George~DC Cavalcanti, Luiz~S Oliveira, Thiago~JM Moura, and Guilherme~V
  Carvalho.
\newblock Combining diversity measures for ensemble pruning.
\newblock \emph{Pattern Recognition Letters}, 74:\penalty0 38--45, 2016.

\bibitem[Demir and Yener(2005)]{demir2005automated}
Cigdem Demir and B{\"u}lent Yener.
\newblock Automated cancer diagnosis based on histopathological images: a
  systematic survey.
\newblock \emph{Rensselaer Polytechnic Institute, Tech. Rep}, 2005.

\bibitem[Limkin et~al.(2017)Limkin, Sun, Dercle, Zacharaki, Robert, Reuz{\'e},
  Schernberg, Paragios, Deutsch, and Fert{\'e}]{limkin2017promises}
EJ~Limkin, Roger Sun, Laurent Dercle, EI~Zacharaki, Charlotte Robert, Sylvain
  Reuz{\'e}, Antoine Schernberg, Nikos Paragios, Eric Deutsch, and Charles
  Fert{\'e}.
\newblock Promises and challenges for the implementation of computational
  medical imaging (radiomics) in oncology.
\newblock \emph{Annals of Oncology}, 28\penalty0 (6):\penalty0 1191--1206,
  2017.

\bibitem[Monkam et~al.(2019)Monkam, Qi, Ma, Gao, Yao, and
  Qian]{monkam2019detection}
Patrice Monkam, Shouliang Qi, He~Ma, Weiming Gao, Yudong Yao, and Wei Qian.
\newblock Detection and classification of pulmonary nodules using convolutional
  neural networks: a survey.
\newblock \emph{IEEE Access}, 7:\penalty0 78075--78091, 2019.

\bibitem[Srinidhi et~al.(2021)Srinidhi, Ciga, and Martel]{srinidhi2021deep}
Chetan~L Srinidhi, Ozan Ciga, and Anne~L Martel.
\newblock Deep neural network models for computational histopathology: A
  survey.
\newblock \emph{Medical Image Analysis}, 67:\penalty0 101813, 2021.

\bibitem[Clark et~al.(2013)Clark, Vendt, Smith, Freymann, Kirby, Koppel, Moore,
  Phillips, Maffitt, Pringle, et~al.]{clark2013cancer}
Kenneth Clark, Bruce Vendt, Kirk Smith, John Freymann, Justin Kirby, Paul
  Koppel, Stephen Moore, Stanley Phillips, David Maffitt, Michael Pringle,
  et~al.
\newblock The cancer imaging archive (tcia): maintaining and operating a public
  information repository.
\newblock \emph{Journal of digital imaging}, 26\penalty0 (6):\penalty0
  1045--1057, 2013.

\bibitem[Tomczak et~al.(2015)Tomczak, Czerwi{\'n}ska, and
  Wiznerowicz]{tomczak2015cancer}
Katarzyna Tomczak, Patrycja Czerwi{\'n}ska, and Maciej Wiznerowicz.
\newblock The cancer genome atlas (tcga): an immeasurable source of knowledge.
\newblock \emph{Contemporary Oncology}, 19\penalty0 (1A):\penalty0 A68, 2015.

\bibitem[D’Amico et~al.(2020)D’Amico, Sicilia, Cordelli, Tronchin, Greco,
  Fiore, Carnevale, Iannello, Ramella, and Soda]{d2020radiomics}
Natascha~Claudia D’Amico, Rosa Sicilia, Ermanno Cordelli, Lorenzo Tronchin,
  Carlo Greco, Michele Fiore, Alessia Carnevale, Giulio Iannello, Sara Ramella,
  and Paolo Soda.
\newblock Radiomics-based prediction of overall survival in lung cancer using
  different volumes-of-interest.
\newblock \emph{Applied Sciences}, 10\penalty0 (18):\penalty0 6425, 2020.

\bibitem[Chaddad et~al.(2015)Chaddad, Zinn, and Colen]{chaddad2015radiomics}
Ahmad Chaddad, Pascal~O Zinn, and Rivka~R Colen.
\newblock Radiomics texture feature extraction for characterizing gbm
  phenotypes using glcm.
\newblock In \emph{2015 IEEE 12th International Symposium on Biomedical Imaging
  (ISBI)}, pages 84--87. IEEE, 2015.

\bibitem[Aggarwal and Agrawal(2012)]{aggarwal2012first}
Namita Aggarwal and RK~Agrawal.
\newblock First and second order statistics features for classification of
  magnetic resonance brain images.
\newblock \emph{Journal of Signal and Information Processing}, 3:\penalty0
  146--153, 2012.

\bibitem[Rashmi et~al.(2020)Rashmi, Prasad, Udupa, and
  Shwetha]{rashmi2020comparative}
R~Rashmi, Keerthana Prasad, Chethana Babu~K Udupa, and V~Shwetha.
\newblock A comparative evaluation of texture features for semantic
  segmentation of breast histopathological images.
\newblock \emph{IEEE Access}, 8:\penalty0 64331--64346, 2020.

\bibitem[Haralick et~al.(1973)Haralick, Shanmugam, and
  Dinstein]{haralick1973textural}
Robert~M Haralick, Karthikeyan Shanmugam, and Its'~Hak Dinstein.
\newblock Textural features for image classification.
\newblock \emph{IEEE Transactions on systems, man, and cybernetics}, pages
  610--621, 1973.

\bibitem[Kr{\'a}l and Lenc(2016)]{kral2016lbp}
Pavel Kr{\'a}l and Ladislav Lenc.
\newblock Lbp features for breast cancer detection.
\newblock In \emph{2016 IEEE International Conference on Image Processing
  (ICIP)}, pages 2643--2647. IEEE, 2016.

\bibitem[Cordelli et~al.(2018)Cordelli, Maulucci, De~Spirito, Rizzi, Pitocco,
  and Soda]{cordelli2018decision}
Ermanno Cordelli, Giuseppe Maulucci, Marco De~Spirito, Alessandro Rizzi, Dario
  Pitocco, and Paolo Soda.
\newblock A decision support system for type 1 diabetes mellitus diagnostics
  based on dual channel analysis of red blood cell membrane fluidity.
\newblock \emph{Computer methods and programs in biomedicine}, 162:\penalty0
  263--271, 2018.

\bibitem[Sicilia et~al.(2019)Sicilia, Cordelli, Merone, Luperto, Papalia,
  Iannello, and Soda]{sicilia2019early}
Rosa Sicilia, Ermanno Cordelli, Mario Merone, Elia Luperto, Rocco Papalia,
  Giulio Iannello, and Paolo Soda.
\newblock Early radiomic experiences in classifying prostate cancer
  aggressiveness using {3D} local binary patterns.
\newblock In \emph{2019 IEEE 32nd International Symposium on Computer-Based
  Medical Systems (CBMS)}, pages 355--360. IEEE, 2019.

\bibitem[Joo et~al.(2021)Joo, Jung, Lee, Park, Kim, Park, and
  Choi]{joo2021stability}
Leehi Joo, Seung~Chai Jung, Hyunna Lee, Seo~Young Park, Minjae Kim, Ji~Eun
  Park, and Keum~Mi Choi.
\newblock {Stability of MRI radiomic features according to various imaging
  parameters in fast scanned T2-FLAIR for acute ischemic stroke patients}.
\newblock \emph{Scientific Reports}, 11\penalty0 (1):\penalty0 1--11, 2021.

\bibitem[Ramella et~al.(2018)Ramella, Fiore, Greco, Cordelli, Sicilia, Merone,
  Molfese, Miele, Cornacchione, Ippolito, et~al.]{ramella2018radiomic}
Sara Ramella, Michele Fiore, Carlo Greco, Ermanno Cordelli, Rosa Sicilia, Mario
  Merone, Elisabetta Molfese, Marianna Miele, Patrizia Cornacchione, Edy
  Ippolito, et~al.
\newblock A radiomic approach for adaptive radiotherapy in non-small cell lung
  cancer patients.
\newblock \emph{PloS one}, 13\penalty0 (11):\penalty0 e0207455, 2018.

\bibitem[Ojala et~al.(1996)Ojala, Pietik{\"a}inen, and
  Harwood]{ojala1996comparative}
Timo Ojala, Matti Pietik{\"a}inen, and David Harwood.
\newblock A comparative study of texture measures with classification based on
  featured distributions.
\newblock \emph{Pattern Recognition}, 29\penalty0 (1):\penalty0 51--59, 1996.

\bibitem[Ojala et~al.(2002)Ojala, Pietikainen, and
  Maenpaa]{ojala2002multiresolution}
Timo Ojala, Matti Pietikainen, and Topi Maenpaa.
\newblock Multiresolution gray-scale and rotation invariant texture
  classification with local binary patterns.
\newblock \emph{IEEE Transactions on pattern analysis and machine
  intelligence}, 24\penalty0 (7):\penalty0 971--987, 2002.

\bibitem[Kuncheva et~al.(2001)Kuncheva, Bezdek, and Duin]{kuncheva2001decision}
Ludmila~I Kuncheva, James~C Bezdek, and Robert~PW Duin.
\newblock Decision templates for multiple classifier fusion: an experimental
  comparison.
\newblock \emph{Pattern Recognition}, 34\penalty0 (2):\penalty0 299--314, 2001.

\bibitem[Kuncheva(2004)]{kuncheva2004combining}
L~Kuncheva.
\newblock Combining pattern classifiers methods and algorithms. john
  wiley\&sons.
\newblock \emph{Inc. Publication, Hoboken}, 2004.

\bibitem[Ho(1995)]{ho1995random}
Tin~Kam Ho.
\newblock Random decision forests.
\newblock In \emph{Proceedings of 3rd international conference on document
  analysis and recognition}, volume~1, pages 278--282. IEEE, 1995.

\bibitem[Buitinck et~al.(2013)Buitinck, Louppe, Blondel, Pedregosa, Mueller,
  Grisel, Niculae, Prettenhofer, Gramfort, Grobler, Layton, VanderPlas, Joly,
  Holt, and Varoquaux]{sklearn_api}
Lars Buitinck, Gilles Louppe, Mathieu Blondel, Fabian Pedregosa, Andreas
  Mueller, Olivier Grisel, Vlad Niculae, Peter Prettenhofer, Alexandre
  Gramfort, Jaques Grobler, Robert Layton, Jake VanderPlas, Arnaud Joly, Brian
  Holt, and Ga{\"{e}}l Varoquaux.
\newblock {API} design for machine learning software: experiences from the
  scikit-learn project.
\newblock In \emph{ECML PKDD Workshop: Languages for Data Mining and Machine
  Learning}, pages 108--122, 2013.

\bibitem[Arcuri and Fraser(2013)]{arcuri2013parameter}
Andrea Arcuri and Gordon Fraser.
\newblock Parameter tuning or default values? an empirical investigation in
  search-based software engineering.
\newblock \emph{Empirical Software Engineering}, 18\penalty0 (3):\penalty0
  594--623, 2013.

\bibitem[Zhang et~al.(2017)Zhang, Liu, Zhang, and Almpanidis]{zhang2017up}
Chongsheng Zhang, Changchang Liu, Xiangliang Zhang, and George Almpanidis.
\newblock An up-to-date comparison of state-of-the-art classification
  algorithms.
\newblock \emph{Expert Systems with Applications}, 82:\penalty0 128--150, 2017.

\bibitem[D'Amico et~al.(2019)D'Amico, Merone, Sicilia, Cordelli, D'Antoni,
  Zanetti, Valbusa, Grossi, Beltramo, Fazzini, et~al.]{d2019tackling}
Natascha~Claudia D'Amico, Mario Merone, Rosa Sicilia, Ermanno Cordelli,
  Federico D'Antoni, Isa~Bossi Zanetti, Giovanni Valbusa, Enzo Grossi,
  Giancarlo Beltramo, Deborah Fazzini, et~al.
\newblock Tackling imbalance radiomics in acoustic neuroma.
\newblock \emph{International Journal of Data Mining and Bioinformatics},
  22\penalty0 (4):\penalty0 365--388, 2019.

\bibitem[Soda et~al.(2021)Soda, D’Amico, Tessadori, Valbusa, Guarrasi,
  Bortolotto, Akbar, Sicilia, Cordelli, Fazzini, et~al.]{soda2021aiforcovid}
Paolo Soda, Natascha~Claudia D’Amico, Jacopo Tessadori, Giovanni Valbusa,
  Valerio Guarrasi, Chandra Bortolotto, Muhammad~Usman Akbar, Rosa Sicilia,
  Ermanno Cordelli, Deborah Fazzini, et~al.
\newblock {AIforCOVID}: predicting the clinical outcomes in patients with
  {COVID-19} applying {AI} to chest-x-rays. an {Italian} multicentre study.
\newblock \emph{Medical image analysis}, 74:\penalty0 102216, 2021.

\bibitem[He et~al.(2016)He, Zhang, Ren, and Sun]{he2016deep}
Kaiming He, Xiangyu Zhang, Shaoqing Ren, and Jian Sun.
\newblock Deep residual learning for image recognition.
\newblock In \emph{Proceedings of the IEEE conference on computer vision and
  pattern recognition}, pages 770--778, 2016.

\bibitem[Szegedy et~al.(2015)Szegedy, Liu, Jia, Sermanet, Reed, Anguelov,
  Erhan, Vanhoucke, and Rabinovich]{szegedy2015going}
Christian Szegedy, Wei Liu, Yangqing Jia, Pierre Sermanet, Scott Reed, Dragomir
  Anguelov, Dumitru Erhan, Vincent Vanhoucke, and Andrew Rabinovich.
\newblock Going deeper with convolutions.
\newblock In \emph{Proceedings of the IEEE conference on computer vision and
  pattern recognition}, pages 1--9, 2015.

\bibitem[Liu et~al.(2021)Liu, Sicilia, Tortora, Cordelli, Nibid, Sabarese,
  Perrone, Fiore, Ramella, and Soda]{liu2021exploring}
Charles~Z Liu, Rosa Sicilia, Matteo Tortora, Ermanno Cordelli, Lorenzo Nibid,
  Giovanna Sabarese, Giuseppe Perrone, Michele Fiore, Sara Ramella, and Paolo
  Soda.
\newblock Exploring deep pathomics in lung cancer.
\newblock In \emph{2021 IEEE 34th International Symposium on Computer-Based
  Medical Systems (CBMS)}, pages 407--412. IEEE, 2021.

\bibitem[Paszke et~al.(2017)Paszke, Gross, Chintala, Chanan, Yang, DeVito, Lin,
  Desmaison, Antiga, and Lerer]{paszke2017automatic}
Adam Paszke, Sam Gross, Soumith Chintala, Gregory Chanan, Edward Yang, Zachary
  DeVito, Zeming Lin, Alban Desmaison, Luca Antiga, and Adam Lerer.
\newblock Automatic differentiation in pytorch.
\newblock \emph{NIPS 2017 Autodiff Workshop}, 2017.

\bibitem[Grossmann et~al.(2017)Grossmann, Stringfield, El-Hachem, Bui,
  Velazquez, Parmar, Leijenaar, Haibe-Kains, Lambin, Gillies,
  et~al.]{grossmann2017defining}
Patrick Grossmann, Olya Stringfield, Nehme El-Hachem, Marilyn~M Bui,
  Emmanuel~Rios Velazquez, Chintan Parmar, Ralph~TH Leijenaar, Benjamin
  Haibe-Kains, Philippe Lambin, Robert~J Gillies, et~al.
\newblock Defining the biological basis of radiomic phenotypes in lung cancer.
\newblock \emph{Elife}, 6:\penalty0 e23421, 2017.

\end{thebibliography}
